%% file: AAAI_camera_ready.tex
\def\year{2020}\relax
\documentclass[letterpaper]{report} 
\usepackage{aaai20}  
\usepackage{times}  
\usepackage{helvet} 
\usepackage{courier}  
\usepackage[hyphens]{url}  
\usepackage{graphicx} 
\usepackage{graphicx}  
\usepackage{epsfig}
\usepackage{amsmath,amssymb}
\usepackage{color}
\usepackage{array}
\usepackage{multicol}
\usepackage{multirow}
\usepackage{booktabs}
\usepackage{comment}
\usepackage{siunitx}
\definecolor{renjiao}{RGB}{0,139,139}
\definecolor{ping}{RGB}{200,100,100}
\definecolor{steve}{RGB}{150,50,150}
\definecolor{red}{RGB}{255,0,0}
\definecolor{blue}{RGB}{0,0,255}
\title{Leveraging Multi-view Image Sets for Unsupervised Intrinsic Image Decomposition and Highlight Separation}
\author{Renjiao Yi,\textsuperscript{\rm 1} Ping Tan,\textsuperscript{\rm 1} Stephen Lin\textsuperscript{\rm 2}\\
\textsuperscript{\rm 1}Simon Fraser University, Burnaby, Canada\\ 
\textsuperscript{\rm 2}Microsoft Research, Beijing, China\\
\{renjiaoy,pingtan\}@sfu.ca, stevelin@microsoft.com 
}
 
\begin{document}

\maketitle

\begin{abstract}
We present an unsupervised approach for factorizing object appearance into highlight, shading, and albedo layers, trained by multi-view real images. To do so, we construct a multi-view dataset by collecting numerous customer product photos online, which exhibit large illumination variations that make them suitable for training of reflectance separation and can facilitate object-level decomposition. 
The main contribution of our approach is a proposed image representation based on local color distributions that allows training to be insensitive to the local misalignments of multi-view images. In addition, we present a new guidance cue for unsupervised training that exploits synergy between highlight separation and intrinsic image decomposition. Over a broad range of objects, our technique is shown to yield state-of-the-art results for both of these tasks.
\end{abstract}

\input{introduction.tex}
\input{related.tex}
\input{overview.tex}
\input{dataset_short.tex}

\input{method.tex}

\input{experiment.tex}
\section{Conclusion}

We proposed an end-to-end network to solve highlight separation and intrinsic image decomposition together. Our network is able to leverage multi-view object-centric image sets, such as our Customer Product Photos Dataset, for unsupervised training via a proposed color distribution loss that is robust to misaligned data. This loss can readily be adapted for other tasks that are sensitive to misalignment.


\small{
\bibliographystyle{aaai}
\bibliography{egbib}
}
\clearpage
\appendix
\chapter{Supplementary Material: \\Leveraging Multi-view Image Sets for Unsupervised Intrinsic Image Decomposition and Highlight Separation}

\section{Ablation studies}

In this section, we present additional ablation results, including further results on an ablation discussed in the main text, as well as other ablation studies.

\subsection{Robustness to misalignment}

\begin{figure}
\centering
\includegraphics[width= \linewidth]{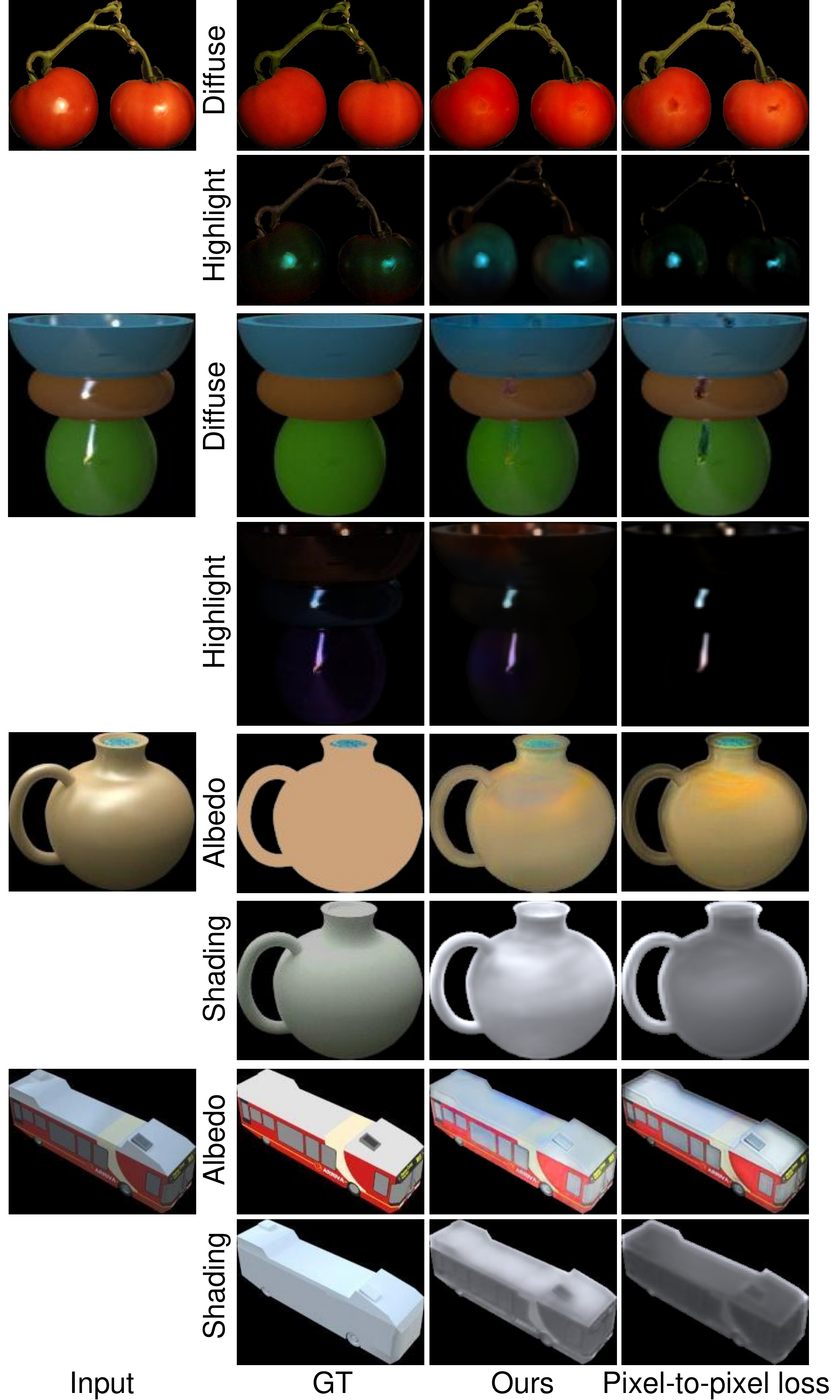}
   \caption{Visual comparisons between our color distribution loss and the pixel-to-pixel low-rank loss in handling misalignment of training images. The top two examples show comparisons on highlight separation, and the bottom two show comparisons on intrinsic image decomposition. 
}
\label{fig:misalignrobust}
\end{figure}

Section \textit{Experiments} of the main text present ablation studies that examine the robustness of our color distribution loss to misalignment of training images. Quantitative comparisons are given in Table~\ref{table:real} (bottom) in the main text for highlight separation and Table~\ref{table:intrinsicsynthetic} (bottom) in the main text for intrinsic image decomposition. It is shown that the color distribution loss is more robust to misalignment, and here in Figure~\ref{fig:misalignrobust} we display some qualitative results from this comparison. It can be seen that for both highlight separation and intrinsic image decomposition, the pixel-to-pixel low-rank loss is not as effective as our color distribution loss for training our full network. For highlight extraction, many highlights are missed in the results while some are overestimated, which shows that a pixel-to-pixel low-rank loss will suffer from misalignment in training images. For intrinsic image decomposition, problems exist with the pixel-to-pixel low-rank loss as well, giving shading predictions that are often incorrect around edges where misalignments have the greatest impact. Trained on exactly the same data, our network with color distribution losses yields much better results, indicating greater robustness to local misalignments. 

\subsection{Without pretraining on synthetic data}

\begin{figure*}
\centering
\includegraphics[width= 0.9\linewidth]{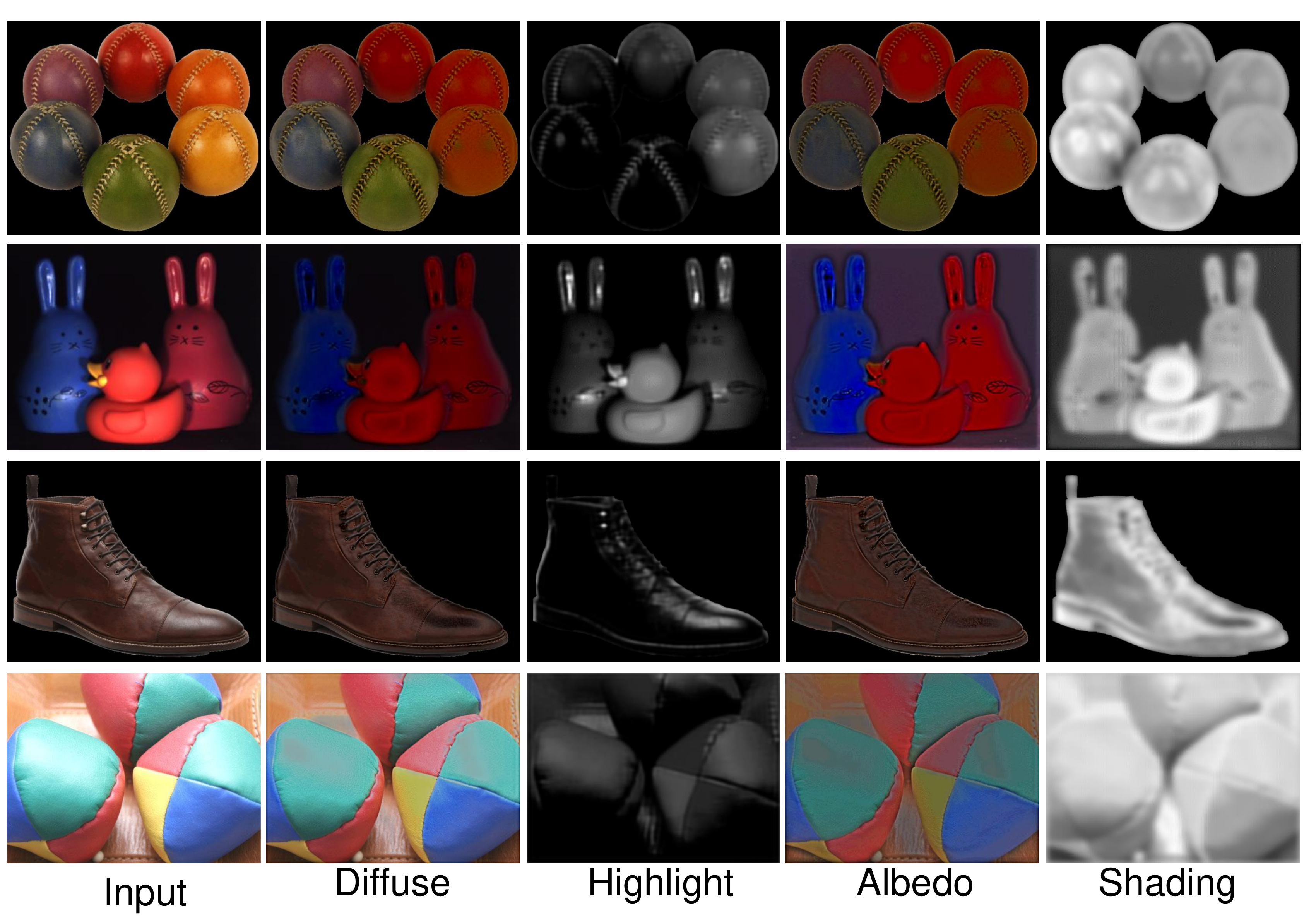}
   \caption{Qualitative results on real images for a fully unsupervised version of our network, without pretraining on synthetic data. 
}
\label{fig:unsupervised}
\end{figure*}

Our model is pretrained on a small amount of synthetic data to bootstrap the unsupervised phases. Here, we examine training the network from scratch with only the unsupervised finetuning. As shown in Figure~\ref{fig:unsupervised}, reasonable highlight extraction and intrinsic image decomposition can be achieved even without pretraining on synthetic data. We evaluated the fully unsupervised network on ShapeNet Intrinsics Dataset and obtained an MSE and DSSIM for highlight extraction of 0.0041 and 0.0227, compared to the leftmost two columns of Table~\ref{table:real} in the main paper. The MSE and DSSIM on real images are 0.0057 and 0.0199, compared to the rightmost two columns of Table~\ref{table:real} in the main paper, which are comparable to previous methods. For intrinsic image decomposition, the MSE and DSSIM are 0.0067 and 0.0527 for albedo, and 0.0059 and 0.0808 for shading, compared to the corresponding values 0.0054 and 0.0436 for albedo, and 0.0045 and 0.0686 for shading in Table~\ref{table:intrinsicsynthetic} of the main paper. This indicates that there is some moderate dropoff without the pretraining on synthetic data, but the performance nevertheless compares well to previous techniques.

\subsection{Using structure-from-motion to align images}

\begin{figure*}
\centering
\includegraphics[width= \linewidth]{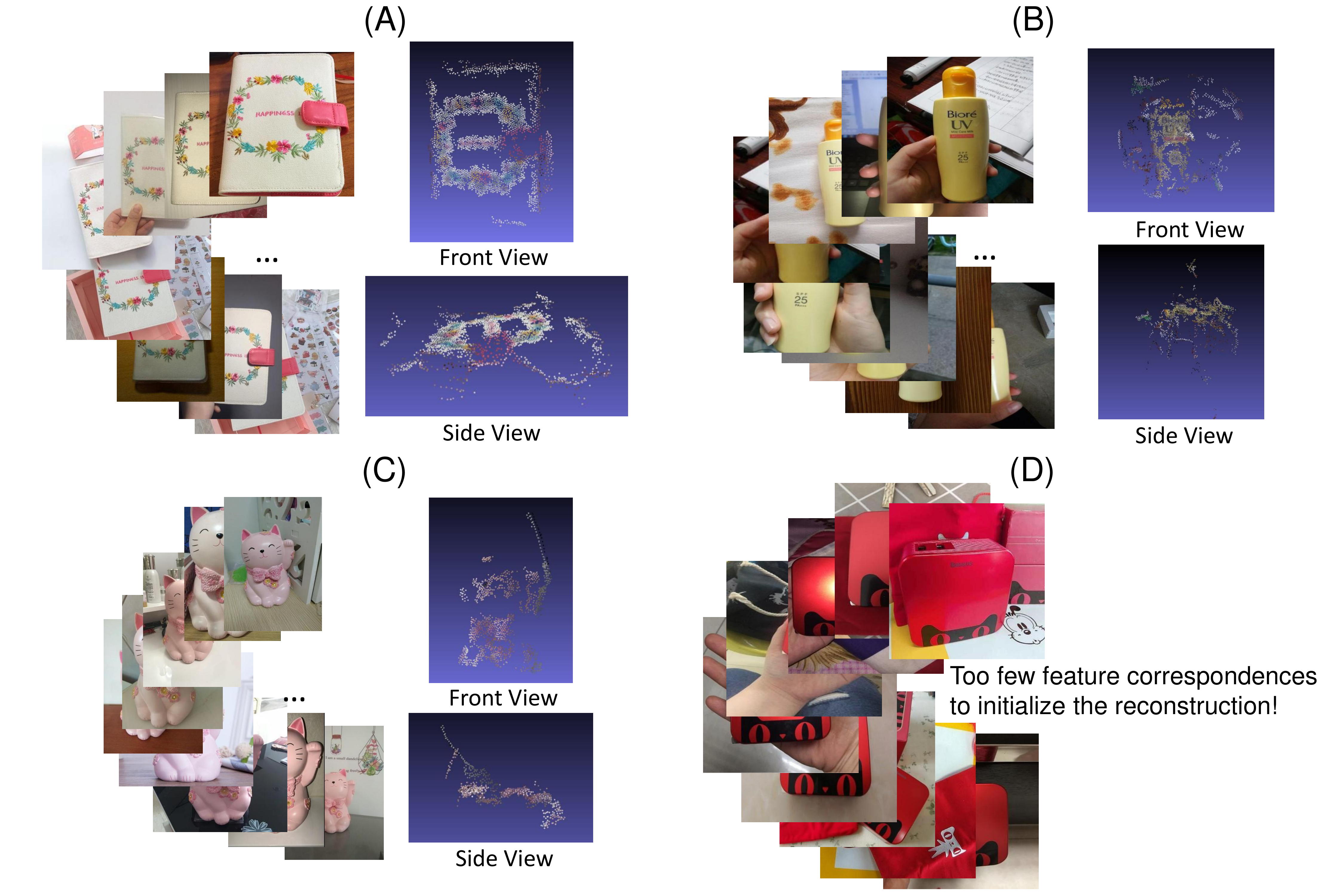}
   \caption{Dense point clouds reconstructed from our multi-view images, using VisualSFM~\cite{wu2011visualsfm} and PMVS2~\cite{furukawa2010accurate}. For each example, selected multi-view images are shown on the left, and reconstructed dense point clouds are shown on the right. 
}
\label{fig:sfm}
\end{figure*}

We explored different alignment methods for our multi-view sets of customer product photos. The most advanced technique is the recent work in \cite{cui2017time}, where they use structure-from-motion to reconstruct the 3D scene and then use it as guidance to align videos of the same scene. Based on this, we implemented a method in which we take 50 multi-view images, reconstruct a sparse point cloud from them using VisualSFM~\cite{wu2011visualsfm}, and then use PMVS2~\cite{furukawa2010accurate} to further reconstruct dense point clouds. However, after applying this structure-from-motion technique to our multi-view customer product photos, we found that many issues exist, as shown in Figure~\ref{fig:sfm} for four examples from our dataset, ordered by increasing difficulty of alignment. Starting from the easiest case (A), a notebook with a minimal highlight layer, the textured regions of the notebook are reconstructed in the point clouds, but textureless regions cannot be reconstructed due to the lack of feature points. However, only the reconstructed regions can be accurately corresponded among the photos. The example in (B) is a common kind of product in our dataset, with some words on a plastic container. It can be seen that only the textured regions can be reconstructed, while some textureless regions are missing, like the cap of the bottle. Another problem is that due to various backgrounds in multi-view images, there are also many background points reconstructed, which adds much noise to the point clouds. Backgrounds are different from image to image and should not be part of the reconstruction, but manually segmenting the foreground in all images is unfeasible. The third example (C) comes from Figure~\ref{fig:dataset} in the main paper. Since it contains little texture for computing feature correspondences, the reconstructed point clouds are very sparse. The fourth example (D) is a metal speaker, which is smooth and glossy. For this kind of object, VisualSFM is unable to provide a result because the feature correspondences are too few. 

In summary, using structure-from-motion to reconstruct an object from our multi-view images faces the following difficulties: the existence of highlights may change the appearance of objects; textureless regions cannot be reconstructed; various backgrounds will lead to much noise; and the number of photos may be too small for certain objects. After much exploration, we found that a combination of WeakAlign~\cite{rocco2018end} and FlowNet2.0~\cite{ilg2017flownet} provides the best alignment results for our customer product photos, but the alignment is not accurate enough to use a pixel-to-pixel loss, as discussed in Section \textit{Ablation Studies} on the first page. 

\section{Customer photos dataset}

In the main paper, due to space limitations, we mention the construction of the dataset very briefly. Here, we describe the steps in more detail: 

{\bf{1. Product selection:}} We manually select product pages containing many customer photos and for which the product does not have multiple versions (e.g., different colors, textures or shapes), so that the product is the same in each photo. We also favor products with an apparent front side, which facilitates alignment.

{\bf{2. Photo downloading:}} We then download customer photos of selected products with batch downloading tools. 

{\bf{3. Rough image alignment:}} For each product, we select one image as the reference and manually segment the object to remove the background. The unconstrained viewpoints and illumination differences among the images makes alignment challenging. We first use WeakAlign~\cite{rocco2018end} to align each of the other images to the segmented reference by an affine transformation. After this global parametric warping, we use FlowNet2.0~\cite{ilg2017flownet} to further align the warped images to the reference. After the transformations of these two steps, the objects in each image will roughly but imperfectly align to the reference. The foreground mask of the reference is used to segment the objects after this alignment. An example of this alignment is shown in the last two rows of Figure~\ref{fig:dataset} of the main text. 

{\bf{4. Data filtering:}} Customer photos exhibit large differences in illumination color as well. To simplify our task, we select photos whose illumination color is similar to that of the reference. This similarity is measured by the difference in median chromaticity. 
We keep only the top $20\%$ of images by this metric. No white balancing is applied, and a gamma of 2.2 is assumed for radiometric calibration. We manually check all the images and remove those with unsuitable content or poor alignment. 

The final Customer Product Photos Dataset consists of 228 products with 10--520 photos for each product. In total, the dataset consists of 9,472 photos. For each product, there is one mask provided for the reference image. The original and aligned images will be made available online upon paper publication.

\section{Additional results on highlight separation}

In addition to the quantitative evaluation and qualitative results shown in the main paper, here we show more qualitative results for highlight separation, with comparisons to previous methods \cite{guo2018single,shi2017learning,shen2013real,yang2010real,tan2005separating}. One bonus of our CNN-based highlight separation method is that it can be used to extract the highlight layers from grayscale images, unlike previous methods which are based on color analysis. 

\subsection{Visual comparisons on ShapeNet Intrinsics Dataset}

\begin{figure*}
\centering
\includegraphics[width= 0.95\linewidth]{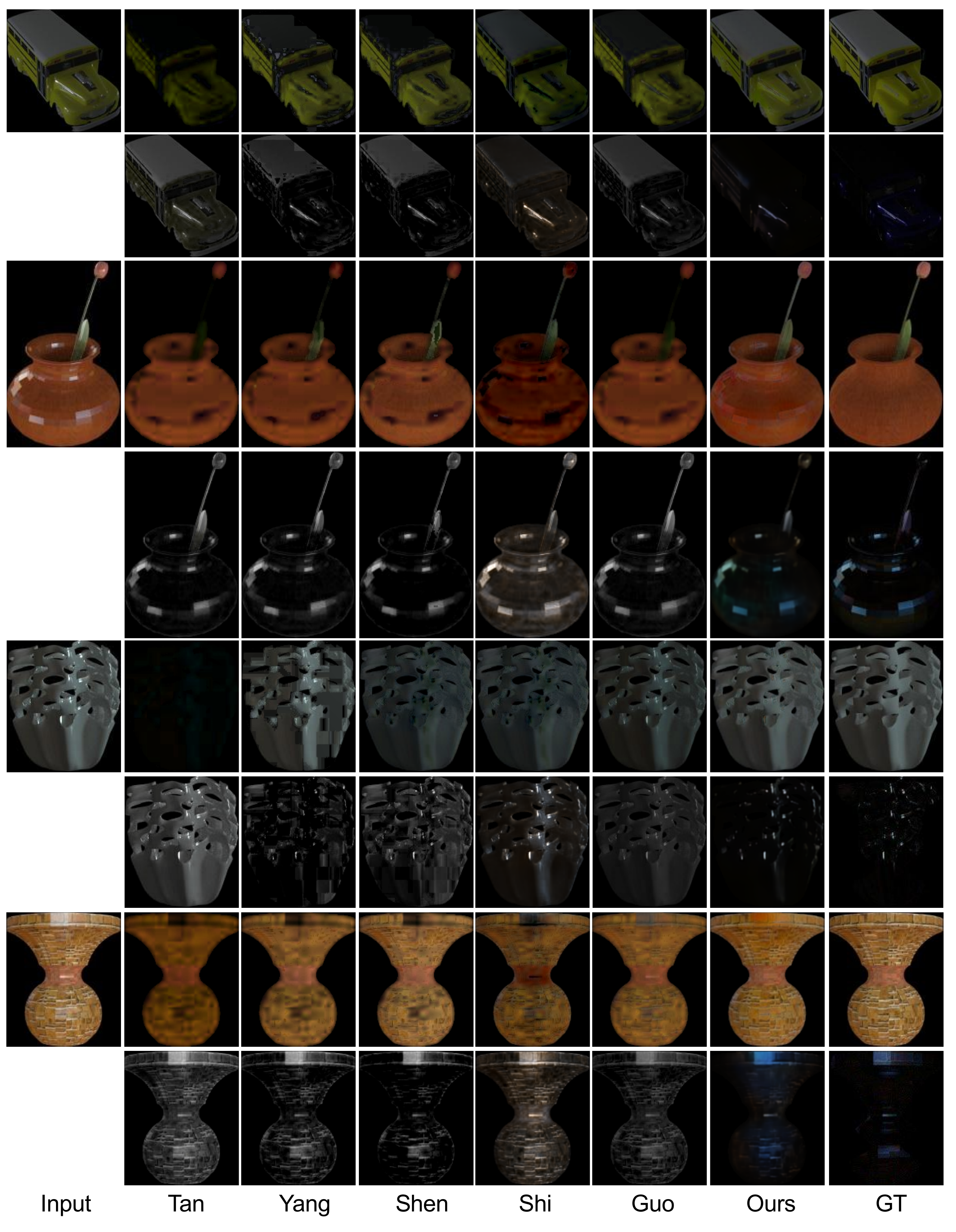}\
   \caption{Qualitative comparisons of highlight separation on ShapeNet Intrinsics Dataset. Tan denotes \cite{tan2005separating}, Yang denotes \cite{yang2010real}, Shen denotes \cite{shen2013real}, Shi denotes \cite{shi2017learning}, Guo denotes \cite{guo2018single}, and GT denotes ground truth separations. For each method, diffuse layers are shown in odd rows and highlight layers are shown in even rows. 
}
\label{fig:synthetichighlihgt1}
\end{figure*}

\begin{figure*}
\centering
\includegraphics[width= 0.86\linewidth]{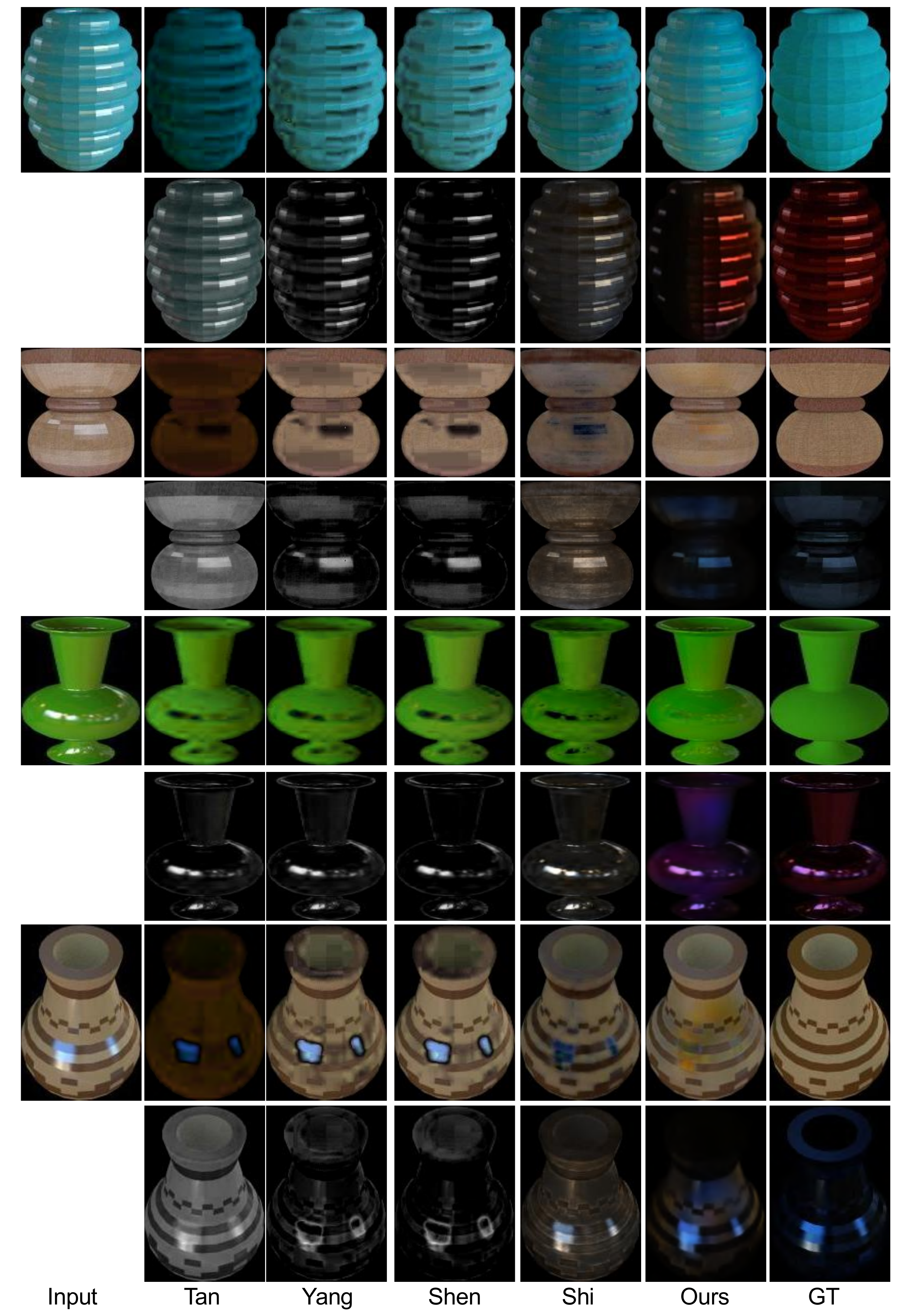}\
   \caption{Qualitative comparisons of highlight separation on ShapeNet Intrinsics Dataset. Tan denotes \cite{tan2005separating}, Yang denotes \cite{yang2010real}, Shen denotes \cite{shen2013real}, Shi denotes \cite{shi2017learning} and GT denotes ground truth separations. For each method, diffuse layers are shown in odd rows and highlight layers are shown in even rows. 
}
\label{fig:synthetichighlihgt2}
\end{figure*}

Additional visual comparisons of highlight separation on ShapeNet Intrinsics Dataset are shown in Figure~\ref{fig:synthetichighlihgt1} and Figure~\ref{fig:synthetichighlihgt2}. Our methods can predict a correct highlight color even when highlight regions are saturated, and the diffuse colors can be recovered correctly. 

\subsection{Visual comparisons on captured real images under lab illumination}

\begin{figure*}
\centering
\includegraphics[width= \linewidth]{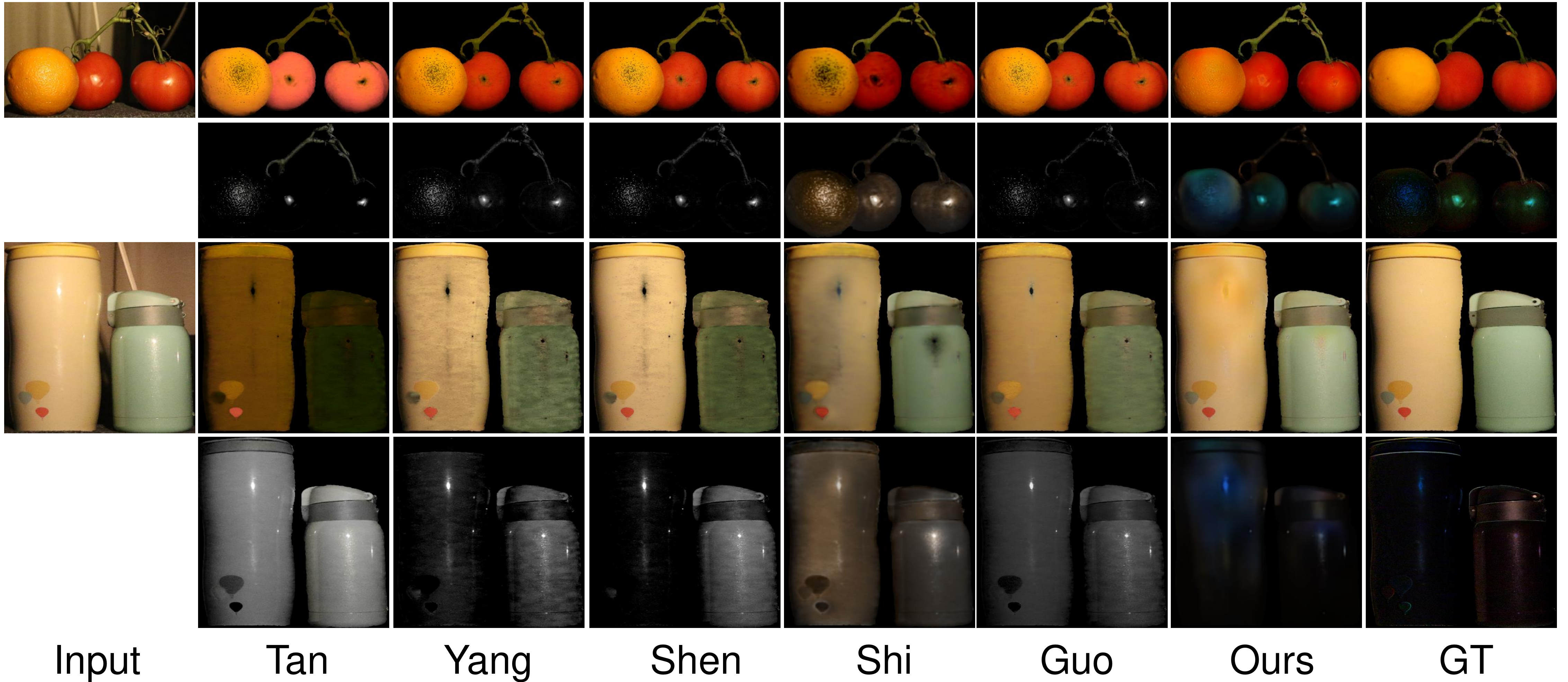}
   \caption{Visual comparisons of highlight extraction on real images. For each example, the top row shows the input image and separated diffuse layers, and the bottom row exhibits the separated highlight layers. Tan denotes \cite{tan2005separating}, Yang denotes \cite{yang2010real}, Shen denotes \cite{shen2013real}, Shi denotes \cite{shi2017learning}, Guo denotes \cite{guo2018single}, and GT denotes ground truth. 
}
\label{fig:highlightrealimages}
\end{figure*}

\begin{figure*}
\centering
\includegraphics[width= \linewidth]{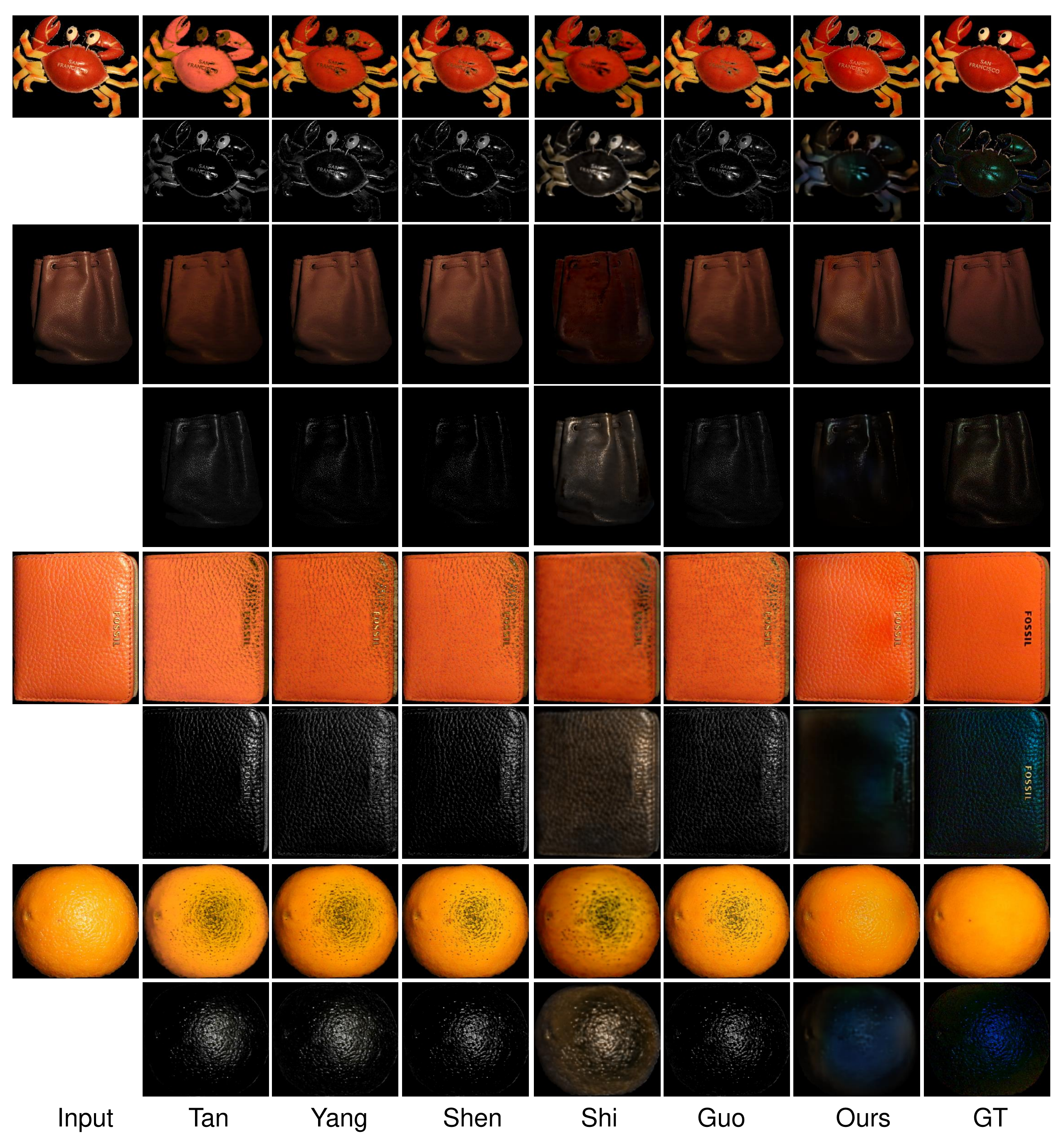}\
   \caption{Qualitative comparisons of highlight separation on captured real images under lab illumination, with ground truth obtained by cross-polarization. Tan denotes \cite{tan2005separating}, Yang denotes \cite{yang2010real}, Shen denotes \cite{shen2013real}, Shi denotes \cite{shi2017learning}, Guo denotes \cite{guo2018single}, and GT denotes ground truth. For each method, diffuse layers are shown in odd rows and highlight layers are shown in even rows. 
}
\label{fig:realhighlight1}
\end{figure*}

For evaluation on real images, since there are no real-image datasets available, we captured a set of real images under lab illumination with ground truth obtained by cross-polarization. Quantitative evaluations on this dataset are shown in the main paper. Here, qualitative comparisons are shown in Figure~\ref{fig:highlightrealimages} and Figure~\ref{fig:realhighlight1}. Most previous methods perform well for images where highlight pixels are not saturated; however, saturation is very common for highlight pixels in real photos, and we find that previous methods tend to overextract the highlight layer and leave some black artifacts on the diffuse layers, as shown in Figure~\ref{fig:highlightrealimages} and the last two examples in Figure~\ref{fig:realhighlight1}.

\subsection{Visual comparisons on real images under natural lighting}

\begin{figure*}
\centering
\includegraphics[width= \linewidth]{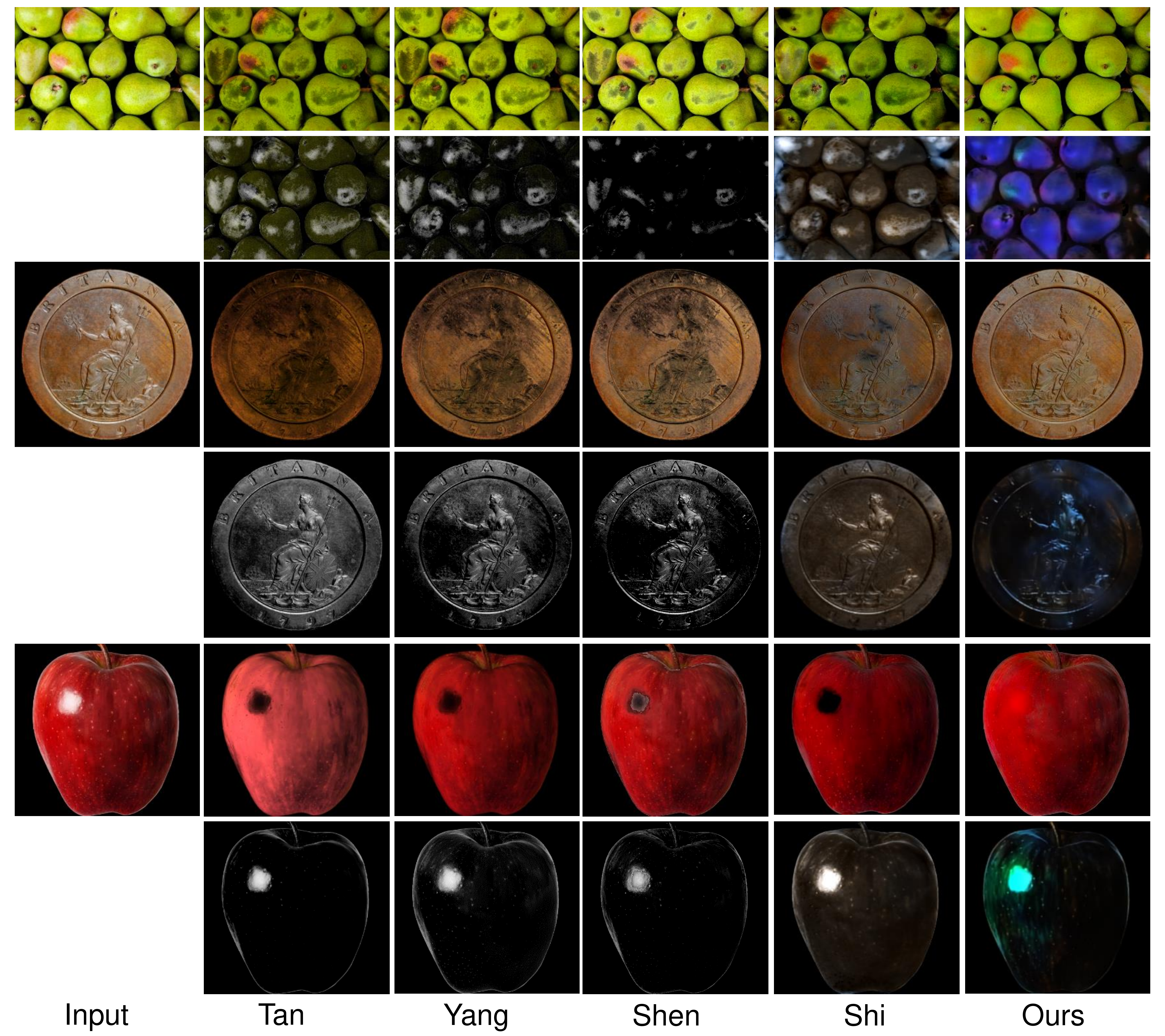}\
   \caption{Qualitative comparisons of highlight separation on real images under natural illumination collected from the Internet, where ground truths are not available. Tan denotes \cite{tan2005separating}, Yang denotes \cite{yang2010real}, Shen denotes \cite{shen2013real} and Shi denotes \cite{shi2017learning}. For each method, diffuse layers are shown in odd rows and highlight layers are shown in even rows. 
}
\label{fig:realhighlightnatural}
\end{figure*}

Besides synthetic images and real images under lab illumination, we also show qualitative comparisons on real images under natural lighting collected from the Internet in Figure~\ref{fig:realhighlightnatural}. We can see that our method succeeds in predicting plausible separations of diffuse layers and highlight layers, for cases with subtle highlights (the first example), glossy metal surfaces (second), and very strong highlights (third). 

\subsection{Highlight separation for grayscale images}

Other than highlight extraction of color images, one advantage of CNN-based methods is that the CNNs trained from color images can also be used on grayscale images, in contrast to conventional methods which rely on color analysis based on the dichromatic model and/or piecewise diffuse colors. 

\begin{figure*}
\centering
\includegraphics[width= \linewidth]{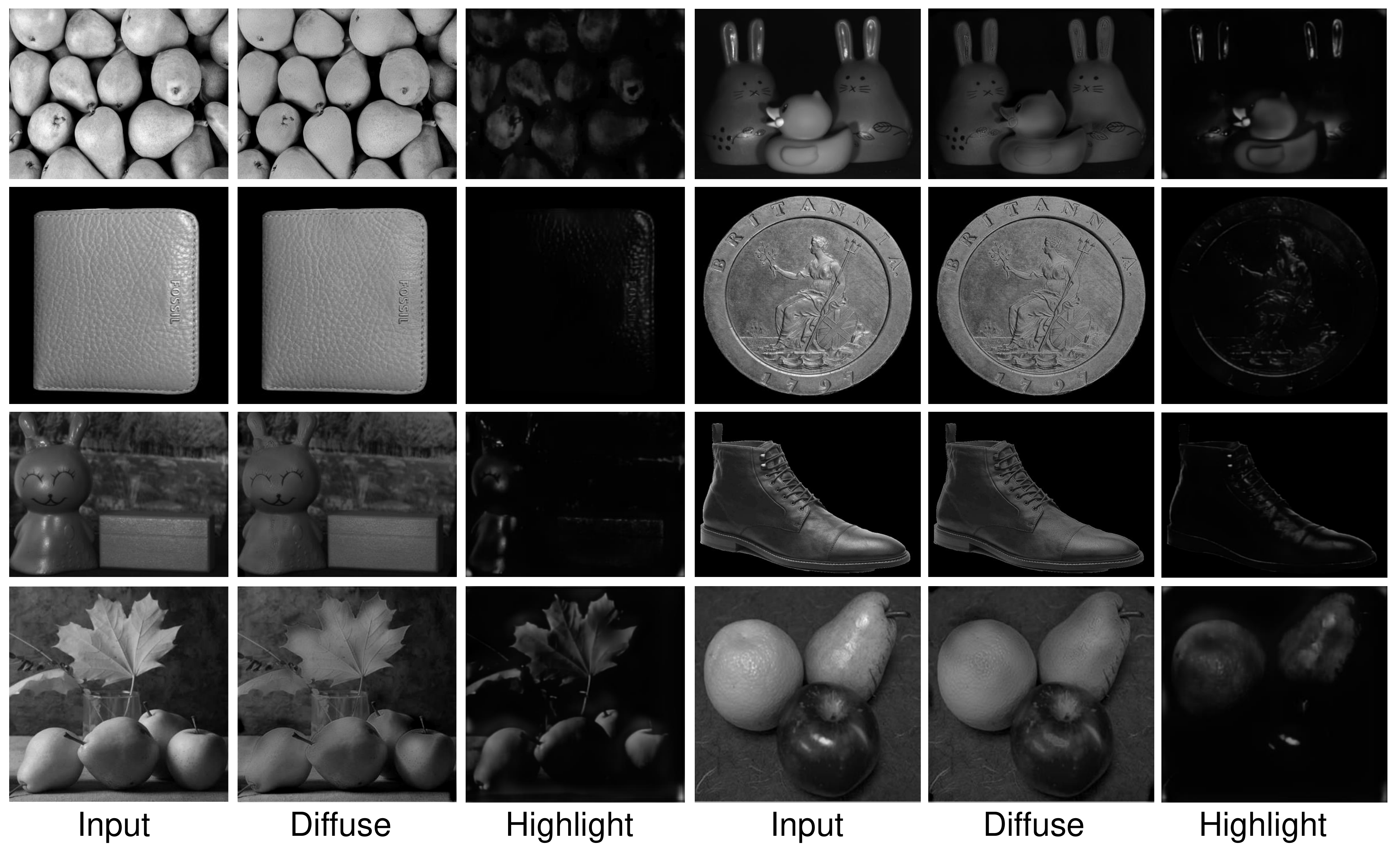}\
   \caption{Qualitative results of highlight separation on grayscale images. 
}
\label{fig:grayscale}
\end{figure*}

For tests on grayscale images, we obtain the predicted highlight in grayscale by averaging its values over the three channels. Subtracting the grayscale highlight layer from the input image gives the diffuse layer. Qualitative results on real images are shown in Figure~\ref{fig:grayscale}. 

\section{Additional results of intrinsic image decomposition}

In this section, additional qualitative comparisons of intrinsic image decomposition on ShapeNet Intrinsics Dataset and the IIW (Intrinsic Images in the Wild) dataset are presented. Additionally, quantitative and qualitative comparisons are given for the MIT intrinsics dataset.

\subsection{MIT intrinsics dataset}

We test our method on the MIT intrinsic image dataset~\cite{grosse2009ground}, which contains real images under white illumination with mostly Lambertian objects. 
For this evaluation, we use S-Net alone, because highlights are merged into the shading in the ground truth decomposition, modeled as $I=A\cdot S$. Since highlights are not correctly represented in this model, the resulting shading contains distortions due to highlight, which we aim to approximate by using S-Net instead of our full system to recover shading.
Despite this less-than-ideal scenario for our method, it still produces reasonable results.

Table~\ref{table:MIT} summarizes the LMSE (an error metric designed specifically for the MIT intrinsics dataset) and MSE (scale-invariant MSE) comparisons. Previous learning based methods, e.g.~\cite{shi2017learning}, generally have problems on this dataset due to the domain shift from synthetic image training to real image testing. Compared to such methods, our S-Net has the advantage of being trainable on multiview sets of real images. SIRFS obtains the best results on this dataset. As noted in previous work~\cite{shi2017learning}, SIRFS is built on priors that match the MIT dataset well (e.g. mostly Lambertian surfaces, white lighting). However, such priors cause SIRFS to be less effective on non-Lambertian objects, as seen in the ShapeNet Intrinsic Dataset experiments.

In the table, we also show results of our S-Net with and without finetuning on the standard MIT training split used by DI~\cite{narihira2015learning}. Due to our network structure, we only use ground truth albedo in training and do not take advantage of ground truth shading. Our shading is computed directly from the additional hard constraint $I=A\cdot S$ once albedo is fixed. From these comparison results, our system demonstrates its advantage of being trainable on a broader range of real images (both fixed- and multi-view).

\begin{figure*}
\centering
\includegraphics[width= \linewidth]{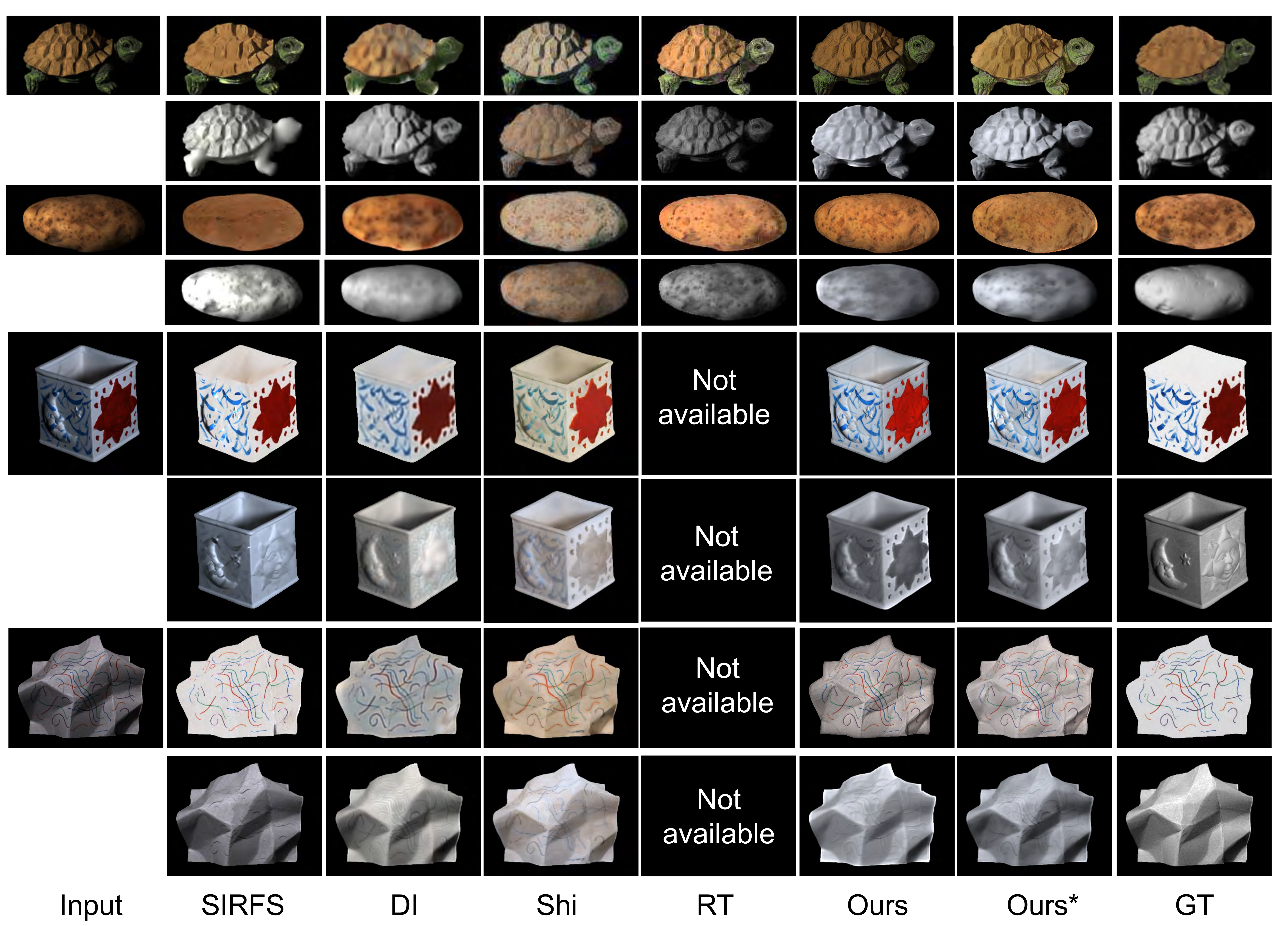}\
   \caption{Visual comparisons of intrinsic image results on the MIT intrinsics dataset. SIRFS denotes \cite{barron2013intrinsic}, DI denotes \cite{narihira2015direct}, Shi denotes \cite{shi2017learning} and RT denotes \cite{baslamisli2018cnn}. Ours denotes our S-Net without finetuning on MIT, and Ours* denotes our S-Net after finetuning on MIT. Since RT\cite{baslamisli2018cnn} does not have code released, we use the images from their paper for the first two data, and their results for the last two data are not available. 
}
\label{fig:MIT}
\end{figure*}

Qualitative comparison examples are shown in Figure~\ref{fig:MIT}. The recovered albedo maps from our method have the highest resolution and most texture detail, while other learning-based methods tend to obtain blurred results.

\begin{table*}[t]
\centering 
\scalebox{1}{
\begin{tabular}{ccc cc ccc} 
\hline 
&&&\multicolumn{2}{c}{LMSE}&&\multicolumn{2}{c}{MSE}\\
\cline{4-5} \cline{7-8}
Method&Training set&&albedo&shading&&albedo&shading\\
\hline
SIRFS\cite{barron2013intrinsic}&MIT&&\textcolor{red}{0.0416}&\textcolor{red}{0.0168}&&\textcolor{blue}{0.0147}&\textcolor{red}{0.0083}\\
DI\cite{narihira2015direct}&MIT+ST&&0.0585&0.0295&&0.0277&0.0154\\
Shi\cite{shi2017learning}&SN&&0.0752&0.0318&&0.0468&0.0194\\
RT\cite{baslamisli2018cnn}&SN2&&0.0652&0.0746&&\textcolor{red}{0.0128}&\textcolor{blue}{0.0107}\\
Ours&SN+CP&&0.0520&0.0416&&0.0365&0.0272\\
Ours*&SN+CP+MIT&&\textcolor{blue}{0.0476}&\textcolor{blue}{0.0284}&&0.0274&0.0145\\
\hline
\end{tabular}
} 
\caption{Intrinsic decomposition on the MIT intrinsics dataset. For the training set, ST denotes ResynthSintel dataset\cite{narihira2015direct}, SN denotes ShapeNet intrinsics dataset, SN2 denotes a similar synthetic dataset created by \cite{baslamisli2018cnn} rendered from ShapeNet models and CP denotes our Customer Photos Dataset. * indicates finetuning on the MIT split used in DI. \label{table:MIT}} 
\end{table*}

\subsection{IIW dataset}

\begin{figure*}
\centering
\includegraphics[width= \linewidth]{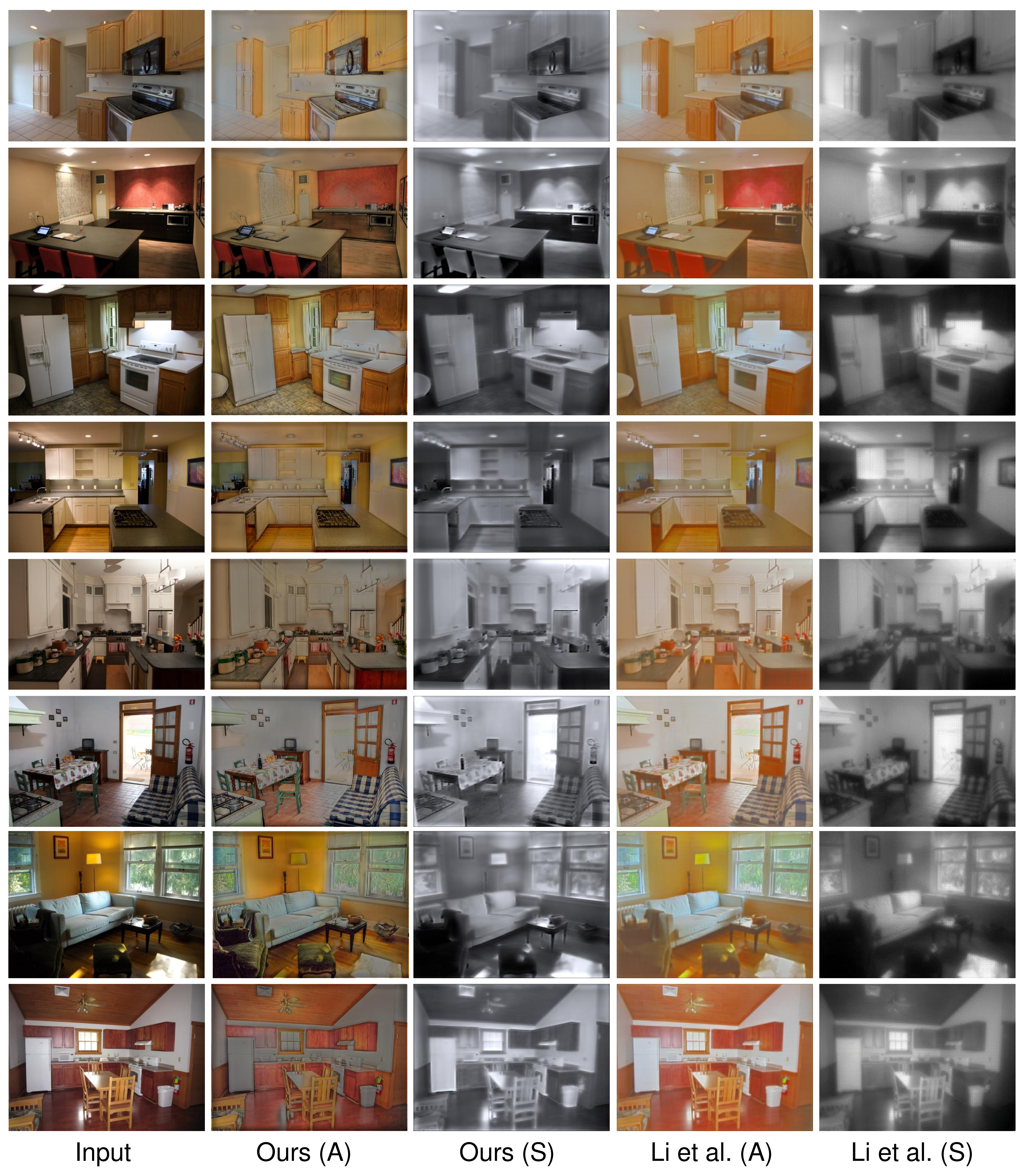}\
   \caption{Qualitative comparisons to Li and Snavely~\cite{li2018learning} on scene images from the IIW dataset. (A) denotes reflectance/albedo results, and (S) denotes shading results.  
}
\label{fig:iiwsupple}
\end{figure*}

\begin{figure*}
\centering
\includegraphics[width= \linewidth]{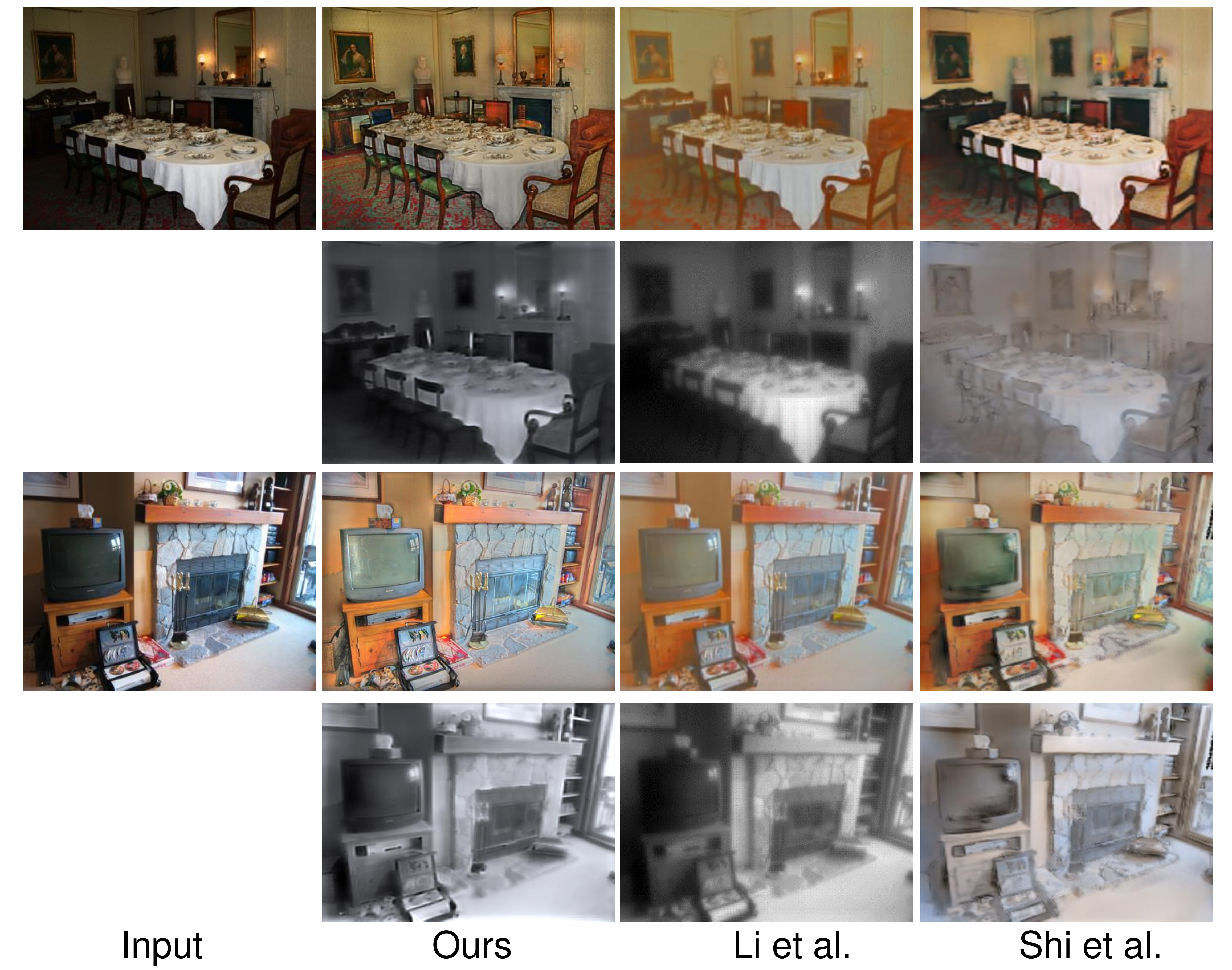}\
   \caption{Qualitative comparisons to Li and Snavely~\cite{li2018learning} and Shi et al.~\cite{shi2019diligent} on scene images from the IIW dataset. The odd rows are input image and albedo results, and the even rows are shading results. 
}
\label{fig:iiw}
\end{figure*}

We also evaluate our network on the Intrinsic Images in the Wild~\cite{bell2014intrinsic} testing set. As mentioned in the main paper, our method is targeted on object-centric images rather than scene images. The WHDR (the Weighted Human Disagreement Rate) evaluation is in Table \ref{table:iiw}. Quantitatively, our results are not as good as methods trained on scene images, but compare favorably to Shi et al.~\cite{shi2017learning}, which is also trained on object-centric images. As shown in Figure~\ref{fig:iiwsupple}, our network also generates qualitative results comparable to Li and Snavely~\cite{li2018learning} trained on scene photos. Comparisons to Shi et al.~\cite{shi2017learning} are shown in Figure~\ref{fig:iiw}, which show that even though our method and Shi et al.~\cite{shi2017learning} are both trained by object-centric images, our method generalizes better on scene images, due to the benefits of real training images.  

\begin{table}[t]
\centering 
\scalebox{0.9}{
\begin{tabular}{cccccc} 
\hline 
Method&DI&Shi&Zhou&Li&Ours\\
\hline
Training set&ST&SN&IIW&BT&CP\\
WHDR\%&37.3&59.4&19.9&20.3&51.1\\
\hline
\end{tabular}
}
\caption{{\bf{Results on the IIW test set}}. Lower is better for the Weighted Human Disagreement Rate (WHDR). ST, BT denote Sintel\cite{butler2012naturalistic} and BigTime\cite{li2018learning} respectively, which are scene datasets. SN and CP denote ShapeNet Intrinsics\cite{shi2017learning} and our Customer Photos datasets respectively, which are object-centric datasets. We evaluate our method and several previous methods, namely DI\cite{narihira2015direct}, Shi\cite{shi2017learning}, Zhou\cite{zhou2015learning} and Li\cite{li2018learning}, on this test set. \label{table:iiw}} 
\end{table}

\subsection{Visual comparisons on ShapeNet dataset}

\begin{figure*}
\centering
\includegraphics[width= 0.9\linewidth]{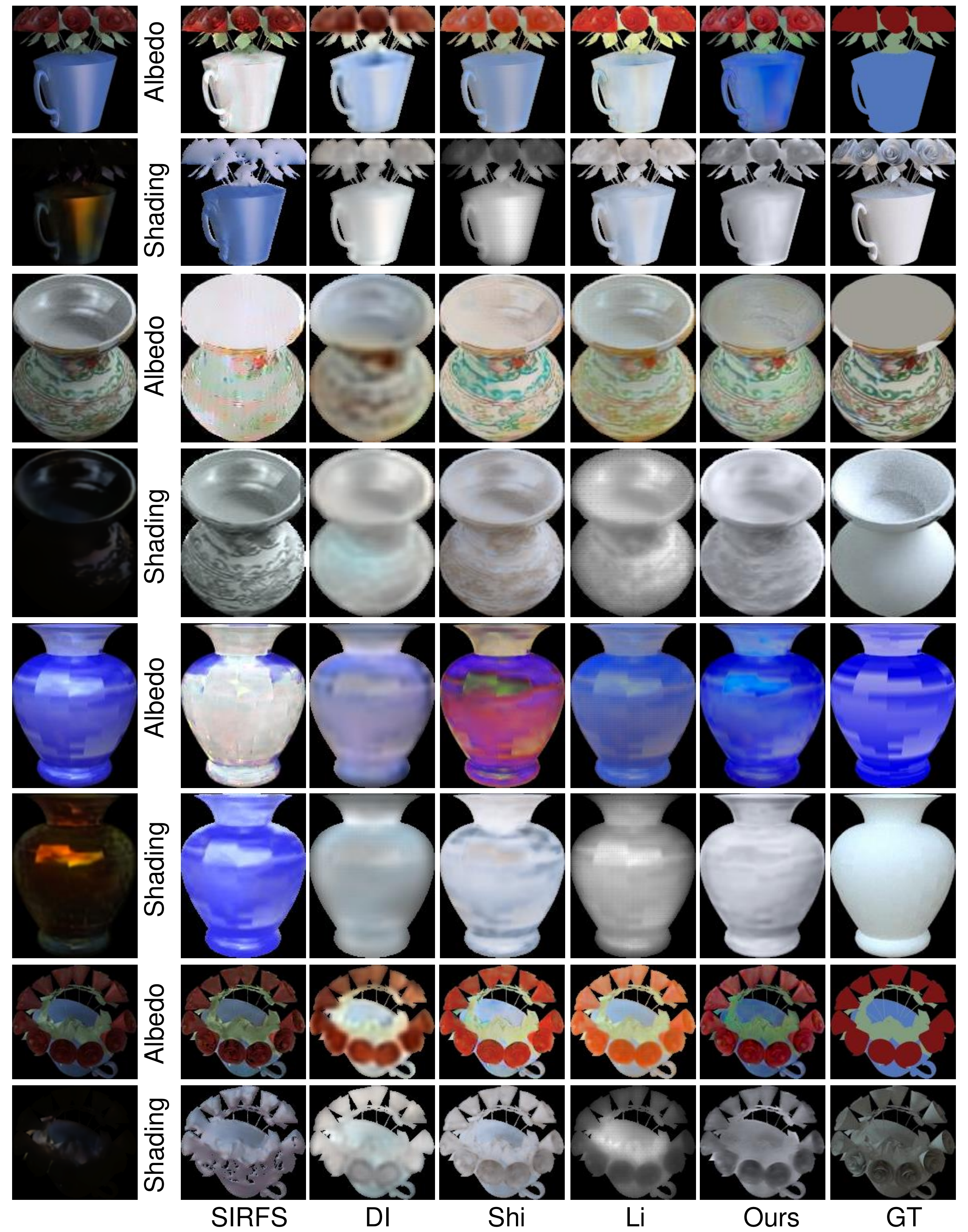}\
   \caption{Additional visual comparisons on ShapeNet Intrinsics Dataset. For the first column, input images are shown at odd rows, our separated highlight layers are shown at even rows. SIRFS denotes \cite{barron2015shape}, DI denotes \cite{narihira2015direct}, Shi denotes \cite{shi2017learning}, and Li denotes \cite{li2018learning}.
}
\label{fig:syntheticintrinsicsupple}
\end{figure*}

Other than the quantitative evaluations in the main paper, qualitative comparisons of intrinsic image decomposition on ShapeNet Intrinsics Dataset are shown in Figure~ref{fig:shapenetintrinsic} and Figure~\ref{fig:syntheticintrinsicsupple}. Here, our full net is used on these non-Lambertian objects, where input images are separated into highlight, albedo and shading layers. All three predicted layers are shown in the figure. By considering the additive highlight layer, albedos generated by our method have much less artifacts on highlight regions. 

\begin{figure*}
\centering
\includegraphics[width= \linewidth]{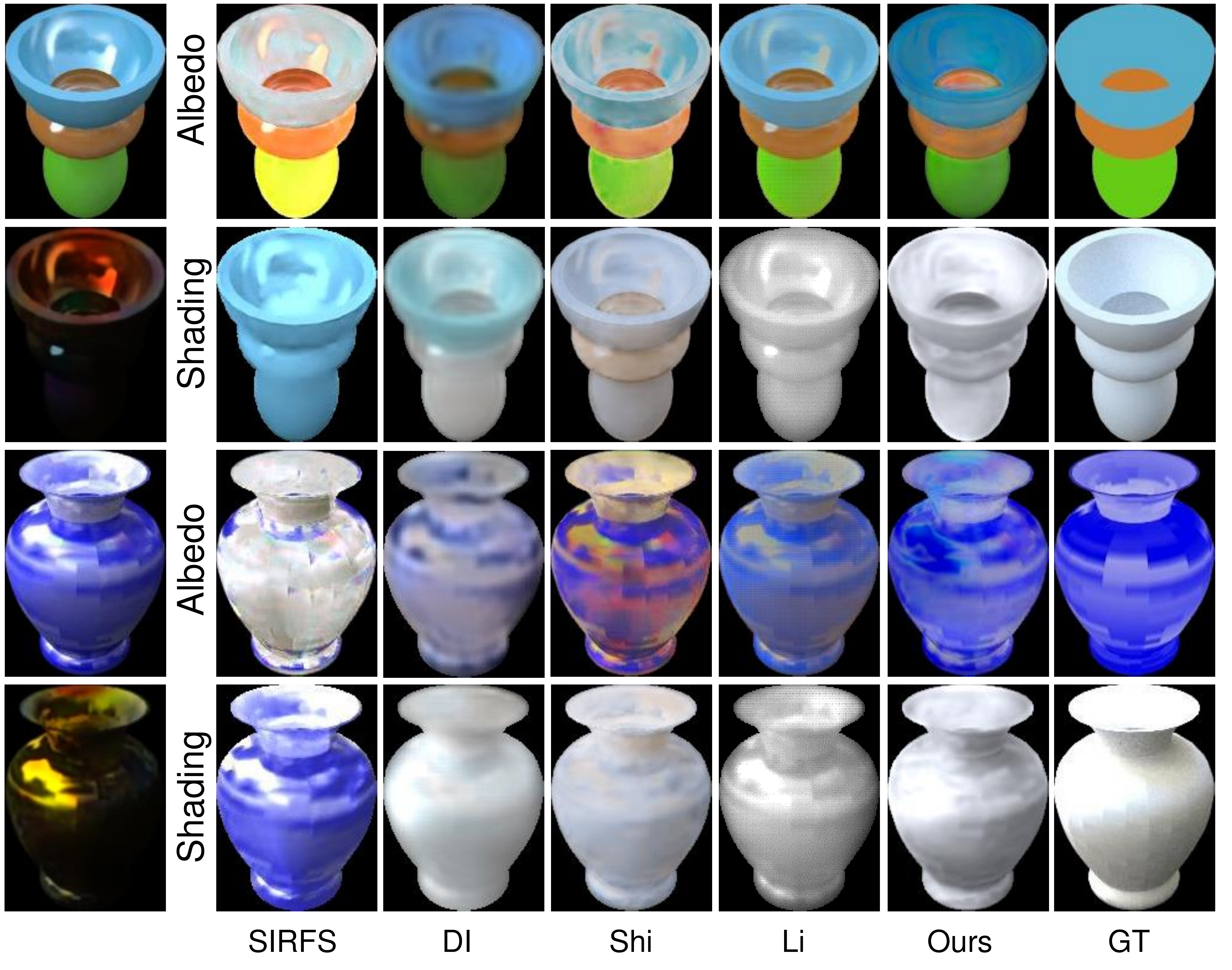}
   \caption{Visual comparisons of intrinsic image decomposition on testing data from the ShapeNet Intrinsic Dataset. For the first column, odd rows show input images and even rows show our separated highlights. SIRFS denotes \cite{barron2015shape}, DI denotes \cite{narihira2015direct}, Shi denotes \cite{shi2017learning}, and Li denotes \cite{li2018learning}. 
    } 
\label{fig:shapenetintrinsic}
\end{figure*}

\section{Evaluation of end-to-end separations}\label{end2end}

\begin{figure*}
\centering
\includegraphics[width= \linewidth]{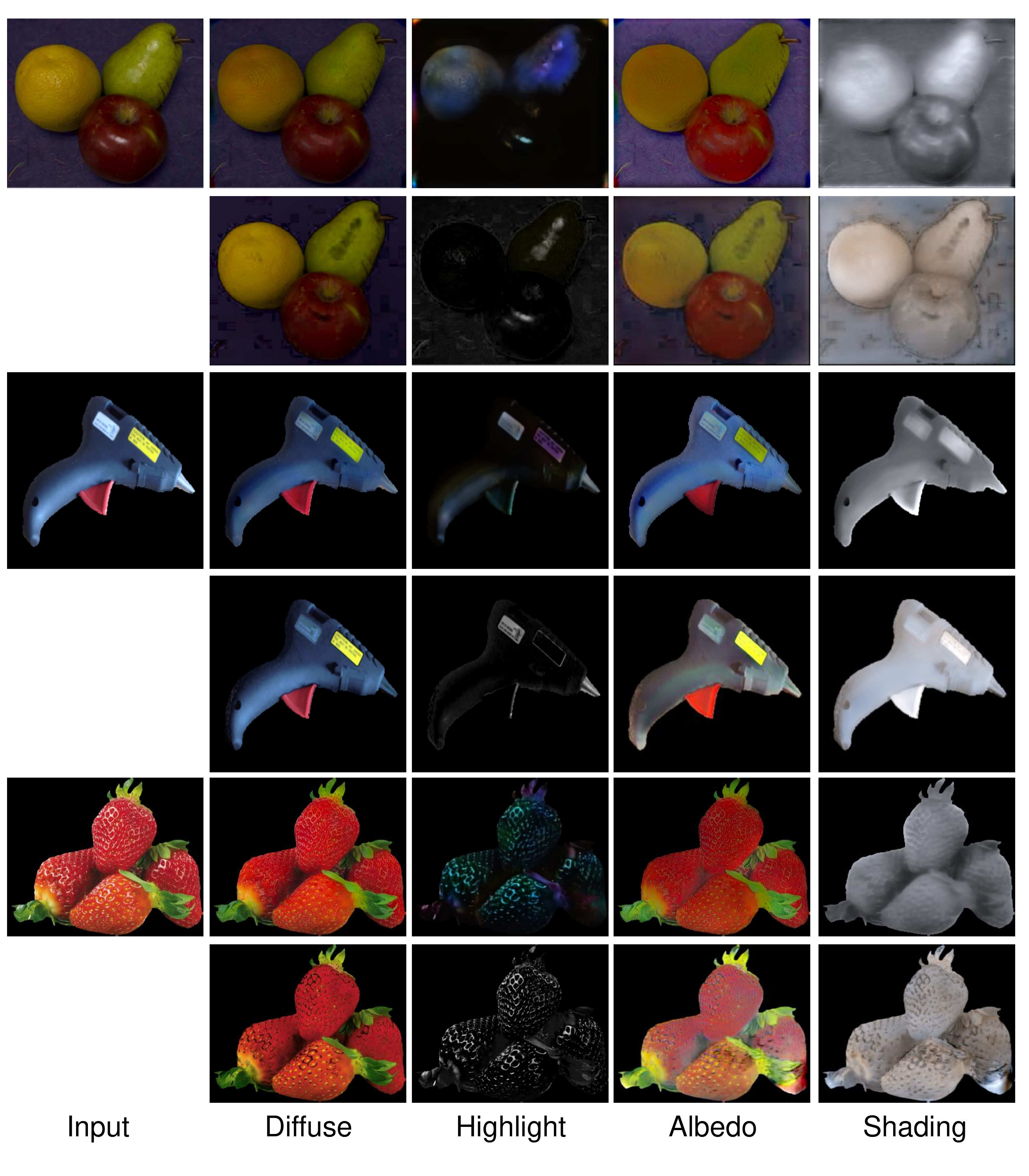}\
   \caption{Qualitative comparisons on real images. We compare our end-to-end separation of highlight, diffuse, albedo and shading layers to the combination of Yang et al.~\cite{yang2010real} for highlight separation and Shi et al.~\cite{shi2017learning} for intrinsic image decomposition, which have the second best performance in quantitative evaluations. The odd rows are our results, and even rows are results of Yang et al.~\cite{yang2010real} and Shi et al.~\cite{shi2017learning}. 
}
\label{fig:qualitativesupple}
\end{figure*}

To evaluate the performance of our end-to-end network, we separate real images into highlight, diffuse, albedo, and shading layers all at once, assuming the image formation model $I=H+A\cdot S$. For comparison, we combine the methods by Yang et al.~\cite{yang2010real} for highlight separation and Shi et al.~\cite{shi2017learning} for intrinsic image decomposition, which have state-of-the-art performance for these tasks. The highlight in the input image is first computed by the method by Yang et al.~\cite{yang2010real} and separated from the input image. The remaining diffuse image is then decomposed into albedo and shading by the method of Shi et al.~\cite{shi2017learning}. As shown in Figure~\ref{fig:qualitativesupple}, our method shows better performance than the combination of Yang et al.~\cite{yang2010real} and Shi et al.~\cite{shi2017learning}, and performs well even on scenes with strong highlights and complicated textures. 

\section{Training details}

\begin{figure*}
\centering
\includegraphics[width=0.9 \linewidth]{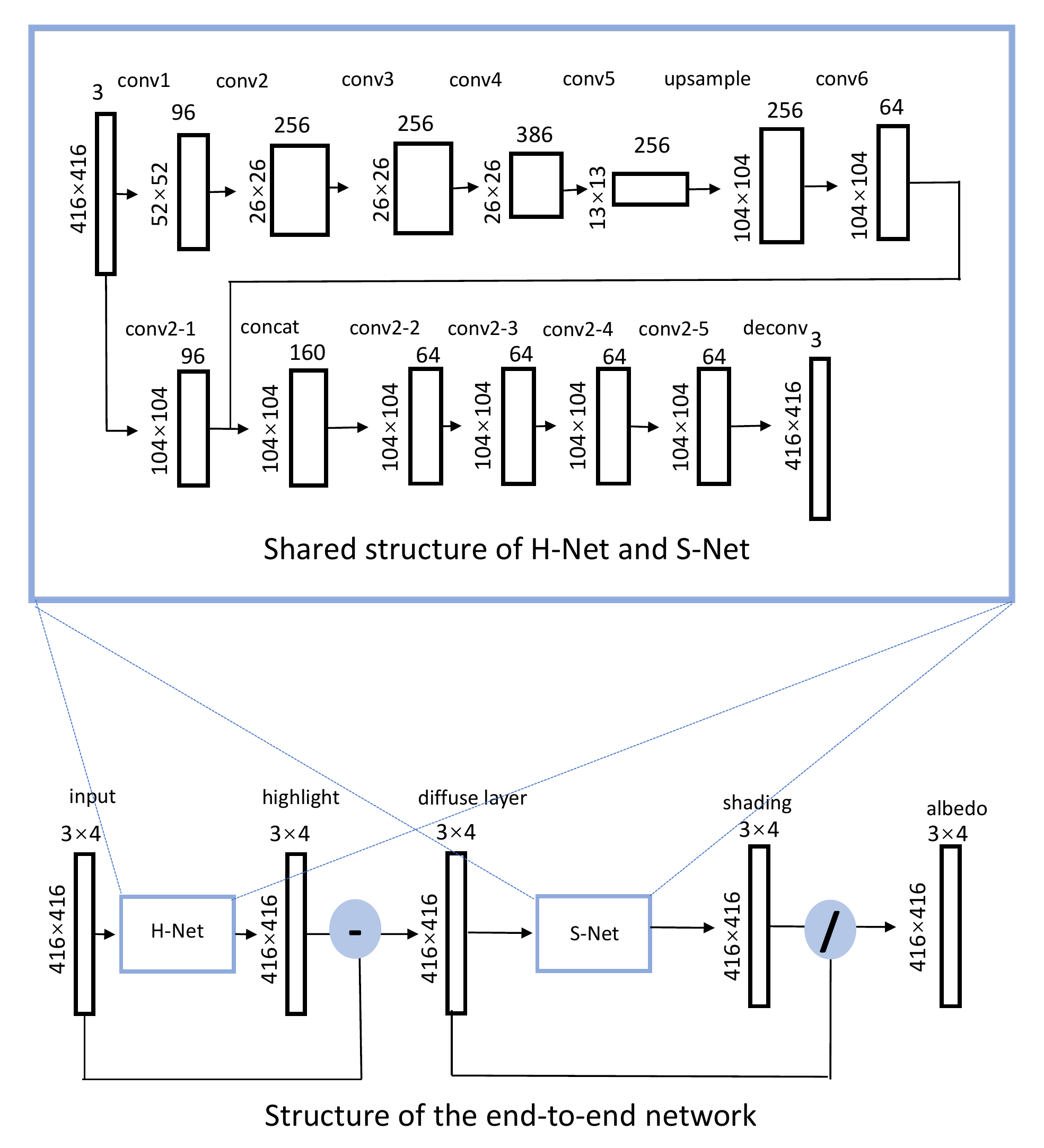}\
   \caption{The structure of our end-to-end network which separates highlight, albedo and shading jointly from a image. The structures of H-net and S-Net are shared, which is adopted from \cite{narihira2015direct} with added batch normalization layers after convolution layers. }
\label{fig:structure}
\end{figure*}

The structure of the end-to-end network is shown in Figure~\ref{fig:structure}. Network structures of H-Net and S-Net are shared, which is a encoder-decoder adopted from \cite{narihira2015direct} with an added batch normalization layer after each convolution layer. In pretraining, the batch size is 32, and the network is pretrained for 1 epoch to provide a reasonable initialization for finetuning. In the unsupervised finetuning phase by low-rank loss, at each batch, 4 objects are randomly selected and 8 images of each object are randomly selected, and then the network is finetuned for 10 epochs. In the joint finetuning phase, the network is finetuned jointly until convergence, which is about 5 epochs. 

\end{document}

%% file: introduction.tex
\section{Introduction}

Separating reflectance layers in an image is an essential step for various image editing and scene understanding tasks. One such layer is composed of highlights, which are mirror-like reflections off the surface of objects. Extracting highlights from an image can be useful for problems such as estimating scene illumination~\cite{lombardi2016reflectance,yi2018faces} and reducing the oily appearance of faces~\cite{li2015simulating}. The other two layers represent shading and albedo. Their separation is commonly known as intrinsic image decomposition, which has been utilized in applications such as shading-based scene reconstruction~\cite{yu2013shading,or-el2015rgbd} and texture replacement in images~\cite{weiss2001deriving,jeon2014intrinsic}.

Factorizing an image into the three reflectance layers is an ill-posed problem that is best solved at present through machine learning. However, obtaining large-scale ground-truth data for training deep neural networks remains a challenge, and this has motivated recent work on developing unsupervised schemes for the reflectance separation problem. The unsupervised techniques that have been presented thus far all take the same approach of training a network on image sequences of a fixed scene under changing illumination~\cite{li2018learning,ma2018single}. With images from such a sequence, these methods guide network training by exploiting the albedo consistency that exists for each scene point throughout the sequence.

\begin{figure}
\centering
\includegraphics[width= 0.95\linewidth]{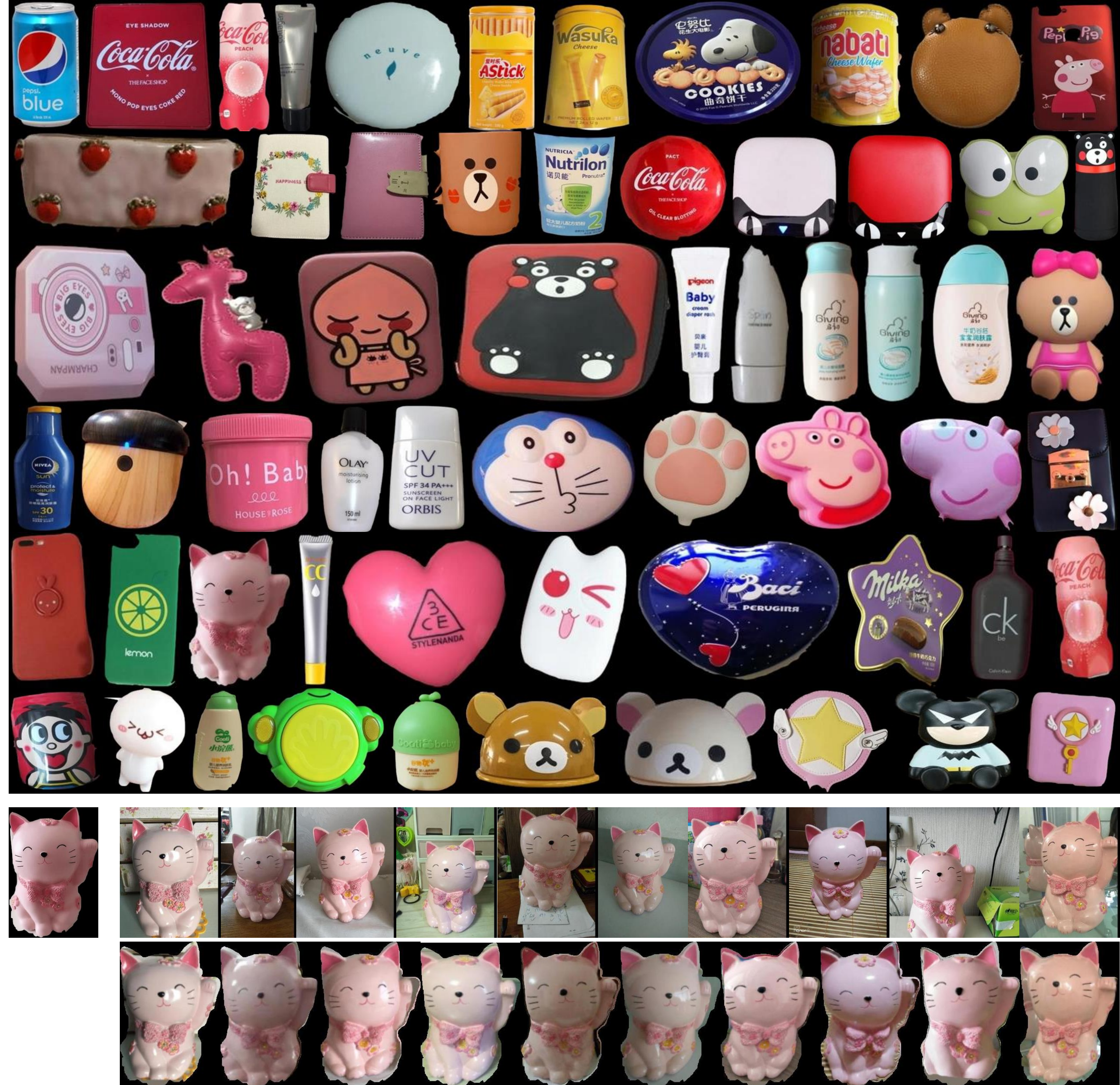}\
   \caption{Selected product photos from the Customer Product Photos Dataset. The products exhibit a wide range of textures, shapes, shadings, and highlight patterns. The second last row shows selected multiview images of the same object, where the leftmost one is the segmented reference image. The last row shows the roughly aligned images. 
}
\label{fig:dataset}
\end{figure}

A benefit of using image sequences of fixed scenes is that the images are perfectly aligned, allowing scene point consistency to be easily utilized. However, there exists an untapped wealth of image data captured of objects from different viewpoints. A prominent example of such data is customer product photos uploaded by consumers to show items they bought. Some example customer photos are shown in Figure~\ref{fig:dataset}. This source of imagery is valuable not just because of its vast quantity online, but also because it provides object-centric data (different from the scene data compiled in~\cite{li2018learning} from webcams) and can promote robustness of factorizations to different object orientations. These images also exhibit a larger variation in illumination conditions and camera settings, which can potentially benefit the trained network. An issue with using such images though is that they are difficult to align accurately, as they vary in viewpoint, lighting and imaging device. Misalignment among the images of an object would lead to violations of scene point consistency on which the existing unsupervised methods are based.

In this paper, we present an unsupervised method for reflectance layer separation using multi-view image sets such as customer product photos. To effectively learn from such data, our system is designed so that its training is relatively insensitive to misalignments. After approximately aligning images with state-of-the-art correspondence estimation techniques~\cite{rocco2018end,ilg2017flownet}, the network transforms the images into a proposed representation based on local color distributions. An important property of this representation is its ability to model detailed local content over an object in a manner that discards fine-scale positional information. With this color distribution based descriptor, unsupervised training becomes possible using consistency constraints between multi-view images of an object.

An additional contribution of this work is a method for further guiding the unsupervised training via a relationship between highlight separation and intrinsic decomposition of shading and albedo. We observe that shading separation becomes less reliable when highlights are present in its input images, due to color distortions caused by different highlight saturation and possibly different illumination color among the images. Our system takes advantage of this through a novel contrastive loss that is defined between shading separation results computed with and without the inclusion of our highlight extraction sub-network. We show that by maximizing this contrastive loss, the shading separation sub-network provides supervision that improves the performance of the highlight extraction sub-network.

With the presented approach, our system produces state-of-the-art results on highlight separation, and yields intrinsic image decomposition accuracy at a level comparable to leading methods. The code and data for this work will be released online upon paper publication.


%% file: related.tex
\section{Related work}

\paragraph{Intrinsic image decomposition}

Previous to the deep-learning approaches of recent years, intrinsic image decomposition was primarily addressed as an optimization problem constrained by various prior assumptions about natural scenes. These priors have been used to classify image derivatives as either albedo or shading change 
\cite{land1971lightness,funt1992recovering}, to prescribe texture coherence \cite{shen2008intrinsic,zhao2012closed}, and to enforce sparsity in the set of albedos \cite{shen2011intrinsic,rother2011recovering}. Decomposition constraints have also been derived using additional input data such as image sequences \cite{weiss2001deriving}, depth measurements \cite{lee2012estimation}, and user input \cite{bousseau2009user}.

These earlier methods have been surpassed in performance by deep neural networks which learn statistical priors from training data. Some of these networks are trained with direct supervision, in which the ground-truth albedo and shading components are provided for each training image~\cite{narihira2015direct,kim2016unified,shi2017learning,baslamisli2018cnn,li2018cgintrinsics}. To obtain ground truth at a large scale for training deep networks, these methods utilize synthetic renderings, which can lead to poor generalization of the networks to real-world scenes. This issue is avoided in several methods by training on sparse annotations of relative reflectance intensity~\cite{bell2014intrinsic} or relative shading \cite{kovacs2017shading} in real images~\cite{zhou2015learning,narihira2015learning,kovacs2017shading,fan2018revisiting}. However, these manual labels provide only weak supervision, and the need for supervision reduces the scalability of the training data.

Most recently, unsupervised methods have been presented in which the training is performed on image sequences taken from fixed-position, time-lapse video with varying illumination~\cite{li2018learning,ma2018single}. In these networks, a major source of guidance for unsupervised training is the temporal consistency of reflectance for static regions within a sequence. The networks are configured so that they can be applied to just a single input image at inference time.


Our proposed system also trains on multiple images in an unsupervised manner and can be applied at test time on single images. Different from the previous fixed-view multi-image techniques~\cite{li2018learning,ma2018single}, our network uses unconstrained multi-view images and deals specifically with misalignment issues that arise in this setting. Such image sequences from unconstrained random views are much easier to obtain than fixed-view images. Moreover, our method additionally separates highlight reflections and introduces a mechanism by which highlight extraction and intrinsic decomposition can mutually benefit each other in unsupervised training.

We note that multiview images have previously been used for intrinsic image decomposition of outdoor scenes~\cite{laffont2013rich,duchene2015multiview}. The decomposition is solved by an inverse rendering approach, where shading is inferred from an approximate multiview stereo reconstruction and an illumination environment estimated given the known sun direction. The multiview images are required to be taken under the same lighting conditions. By contrast, we address a problem where no knowledge about the illumination is given and the lighting can differ from image to image.


\paragraph{Highlight separation}

Similar to intrinsic image decomposition, separation of highlight reflections is an ill-posed problem that has been made tractable through the use of different priors. Among them are priors on piecewise constancy of surface colors~\cite{klinker1988measurement}, smoothness of diffuse~\cite{tan2003highlight}
or specular~\cite{liu2015saturation} reflection, constancy in the maximum diffuse chromaticity~\cite{tan2005separating}, diffuse texture coherence~\cite{tan2006separation}, low diffuse intensity in a color channel~\cite{kim2013specular}, sparsity of highlights~\cite{guo2018single},
and a low-rank representation of diffuse reflection~\cite{guo2018single}.

Instead of crafting priors for highlight extraction by hand, they can be learned in a statistical fashion from images using neural networks. This was first investigated together with intrinsic image decomposition through supervised learning on a large collection of rendered images~\cite{shi2017learning}. An unsupervised approach was later presented for the case of human faces, where a set of images of the same face is aligned using detected facial landmark points, and training guidance is provided by a low-rank constraint on diffuse chromaticity across the aligned images~\cite{yi2018faces}. In \cite{yi2018faces}, face images are easy to align because of mature facial landmark detection techniques; however, their method works poorly on random objects without such landmarks. Thus, we design a much general method to deal with such multi-view images of general objects which are difficult to align accurately. Since misaligned images violate the low-rank property assumed in~\cite{yi2018faces}, we propose a technique that is robust to such local misalignments, thus enabling unsupervised training over a much broader range of objects. Thus, our method is the first unsupervised method using unconstrained images under random illumination, background, and viewpoints. 

%% file: overview.tex
\section{Overview}
We train an end-to-end deep neural network to separate a single image into highlight, albedo/reflectance, and shading layers using the Customer Product Photos Dataset. Compiled from online shopping websites, the dataset contains numerous product photos provided in customer reviews. The photos for a given product are captured under various viewpoints, illumination conditions, and backgrounds. We introduce this dataset in Section \textit{Customer Product Photos Dataset}. 

\begin{figure}
    \centering
   \includegraphics[width= \linewidth]{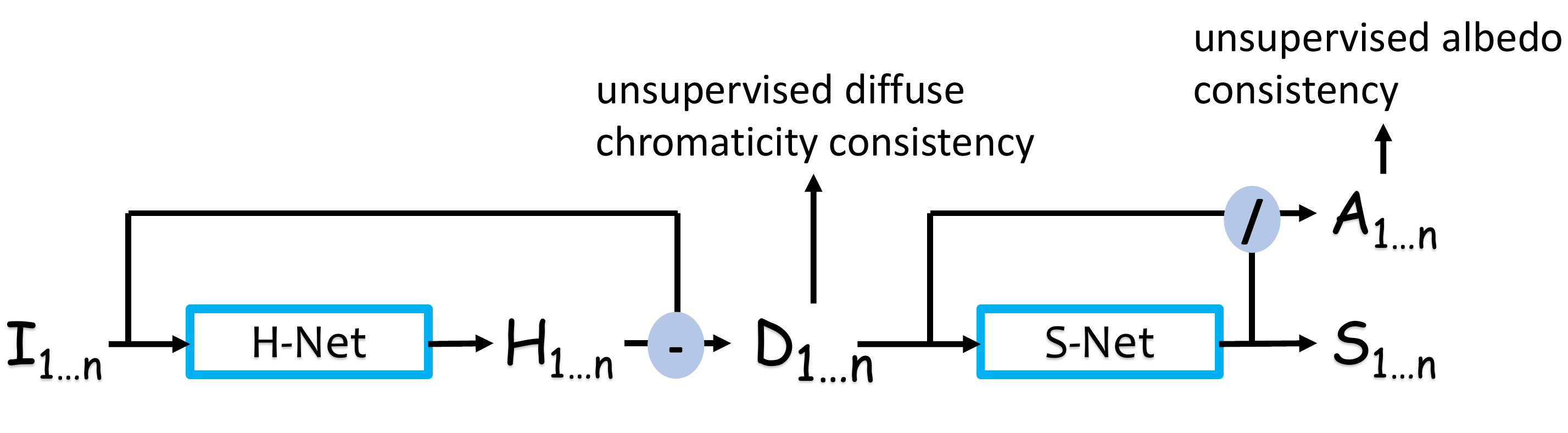}\
    \caption{Network structure. 
}
    \label{fig:pipeline}
\end{figure}

As illustrated in Figure~\ref{fig:pipeline}, our network consists of two subnets: H-Net for decomposing an image into diffuse and highlight layers, and S-Net for additionally decomposing the diffuse layer into albedo and shading layers. Training consists of three phases. First, both H-Net and S-Net are pretrained using a small set of synthetic data from ShapeNet~\cite{shi2017learning}. Each subnet is then finetuned in an unsupervised manner on the Customer Product Photos Dataset using the proposed color distribution loss (Section \textit{Misalignment-robust color distribution loss}), which is robust to misalignments. In the last phase, a novel contrastive loss is used to finetune the whole network end-to-end. The training phases are presented in Section \textit{Our Network}. 

%% file: dataset_short.tex
\section{Customer Product Photos Dataset}\label{customerdataset}

Almost every popular online shopping website includes customer reviews, where customers are often encouraged to upload product photos. For a given product, the customer photos capture it under a various viewpoints, illuminations, and backgrounds. At the same time, the different products cover a large variety of materials and shapes. Collectively, these customer photos capture the complex interaction between different 3D shapes, materials, and illumination, and form a dataset that can be useful for computer vision tasks such as intrinsic image decomposition and multi-view stereo.


Construction of the dataset involved a series of steps consisting of product selection, photo downloading, rough image alignment, and data filtering. Due to limited space, please refer to the supplement for details.

The final Customer Product Photos Dataset consists of 228 products (some shown in Figure~\ref{fig:dataset}) with 10--520 photos for each product. In total, the dataset consists of 9,472 photos. For each product, there is one mask provided for the reference image. The original and aligned images will be made available online upon paper publication.

%% file: method.tex
\section{Our Network} \label{sec:network}
\subsection{Problem formulation}
An input image $I$ comprises an additive combination of a highlight layer $H$ and a diffuse layer $I_d$, where the diffuse layer $I_d$ is a pixelwise product of an albedo/reflectance layer $A$ and a shading layer $S$, i.e.,
\begin{equation}
I=H+I_d = H + A \cdot S.
\label{equ:formation}
\end{equation}
Our problem is to estimate $H, I_d, A, S$ from the input image $I$. We note that this image model differs from the conventional intrinsic image model, $I = A\cdot S$, which omits the additive effects of highlights and thus implicitly assumes object surfaces to be matte~\cite{shi2017learning}.

\subsection{Low-rank loss for unsupervised training}
Most CNN-based methods~\cite{janner2017self,shi2017learning,narihira2015learning} for intrinsic image separation rely completely on ground truth separation results for supervised training. 
As it is difficult to obtain reference ground truth for highlight separation or intrinsic image decomposition on real images, we propose to train our network by unsupervised finetuning on real multiview images after an initial supervised pretraining step with synthetic data from the ShapeNet dataset~\cite{shi2017learning}. This pretraining uses 28,000 out of the 2,443,336 images in the dataset, or about $1.1\%$ of the total, and is intended to provide the network with a good initialization. The finetuning is then intended to adapt the network to the domain of real images, for which ground truth is generally unavailable.

We first assume perfect image alignment in deriving the low-rank loss for unsupervised training. This requirement on alignment will be relaxed in the next subsection.

\paragraph{H-Net}
For training of highlight separation, our network utilizes input consisting of multiple aligned images $I_1, I_2, I_3, \cdots$ of the same object under different lighting. 
According to the image formation model, these images each have a diffuse layer, denoted as $I_{d1}, I_{d2}, I_{d3}, \cdots$. These diffuse layers can differ from each other due to changes in shading that arise from different illumination conditions. To discount this shading variation, we compute the chromaticity maps of these diffuse layers. A chromaticity map $(Ch_r, Ch_g)$ is an intensity-normalized image, where
$$
Ch_r(p) = \frac{R(p)}{R(p)+G(p)+B(p)},
$$
$$
Ch_g(p) = \frac{G(p)}{R(p)+G(p)+B(p)},
$$
at each pixel $p$, with $R(p), G(p), B(p)$ denoting the color values at $p$. 

According to the dichromatic reflectance model \cite{shafer1985using}, the chromaticity of diffuse layers is the chromaticity of the surface albedo multiplied with that of the illumination. 
Assuming a constant illumination color across each image, we discount the effect of illumination chromaticity by matching the median chromaticity of each diffuse image to that of the reference image in each batch. After these normalizations, the set of chromaticity maps should be of low rank if the images are accurately aligned.



The structure of H-Net is adopted from the encoder-decoder network in \cite{narihira2015direct} with an added batch normalization layer after each convolution layer to aid in network convergence. We also examined adding skip connections between the encoder and decoder as done in~\cite{shi2017learning}, but we found them not to be helpful in our network.

\paragraph{S-Net}\label{finetuneintrinsic}
Our S-Net for predicting the shading layer $S$ uses the same network structure as H-Net. The albedo layer $A$ is computed from $S$ at each pixel $p$ according to the image formation model, as
\begin{equation}
\centering
A(p)=I_d(p)/S(p),
\label{equation:albedo}
\end{equation}
once the shading layer is fixed. 

For multiple aligned diffuse images $I_{d1}, I_{d2}, I_{d3},\cdots$ of the same object, their albedo layers $A_1, A_2, A_3, \cdots$ should be the same. Therefore, we can enforce a consistency loss on these different albedo layers for unsupervised training of S-Net.

\paragraph{Low-rank loss}
Our unsupervised training enforces consistency among diffuse chromaticity layers and albedo layers via a low-rank loss. For the case of albedo layers, the low-rank loss can be defined as the second singular value of the matrix $M$ formed by reshaping each albedo image into a vector and stacking the vectors of multiple images~\cite{yi2018faces}. Although consistency could alternatively be enforced through minimizing L1 or L2 differences, e.g. minimizing $|A_1 - A_2|_{1,2}$, the lack of scale invariance of the L1 and L2 losses can lead to degenerate results where $A_1$ and $A_2$ approach zero.
To avoid this problem, the loss function should satisfy the following constraint, $$
\mathcal{L}(A_1,A_2)= \mathcal{L}(\alpha A_1,\alpha A_2),
$$ 
where $\alpha$ is a global scale factor for the whole albedo image.

In order to make the low-rank loss scale-invariant, we use the first singular value to approximate the scale and define a scale-invariant low-rank loss (SILR) as 
\begin{equation}
\begin{split}
& \mathcal{L}_{SILR}=\sigma_2/\sigma_1,\\
&{\frac{\partial \mathcal{L}_{SILR}}{\partial M_{i,j}}}=\frac{\sigma_1*(U_{i,2}\times V_{2,j})-\sigma_2*(U_{i,1}\times V_{1,j})}{\sigma_1^2}.\label{equation:lowrank}
\end{split}
\end{equation}
\noindent{where $\sigma_1$ and $\sigma_2$ are the first two singular value of $M$ computed by SVD decomposition. }
We apply this scale-invariant low-rank loss (SILR) to train both H-Net and S-Net.

\subsection{Misalignment-robust color distribution loss}\label{reorderingloss}
We present a way to relax the requirement of pixel-to-pixel correspondence in the low-rank loss, so that customer photos can be effectively utilized for training. Our observation is that, though precise pixelwise alignment is generally difficult, the state-of-the-art alignment algorithms, e.g. WeakAlign~\cite{rocco2018end} and FlowNet~\cite{ilg2017flownet,dosovitskiy2015flownet}, are mature enough to establish a reasonable approximate alignment. Thus, though some pixels may be misaligned, their correct correspondences are still within a small neighborhood of their estimated locations. This motivates us to develop a local distribution based representation for the low-rank loss.

Suppose we have a predicted albedo layer $A$. We partition it into a grid of $N$ cells. Within each cell, we reorder the pixels by increasing intensity. This is done for each color channel individually, and all the cells for all the color channels are reshaped and concatenated to form a new vector representation for the image. The color distribution loss is then computed as the SILR of these image vectors. In our implementation, we divided 320$\times$320 images into 256 grid cells for all training phases.



\begin{figure}
    \centering
   \includegraphics[width=0.9\linewidth]{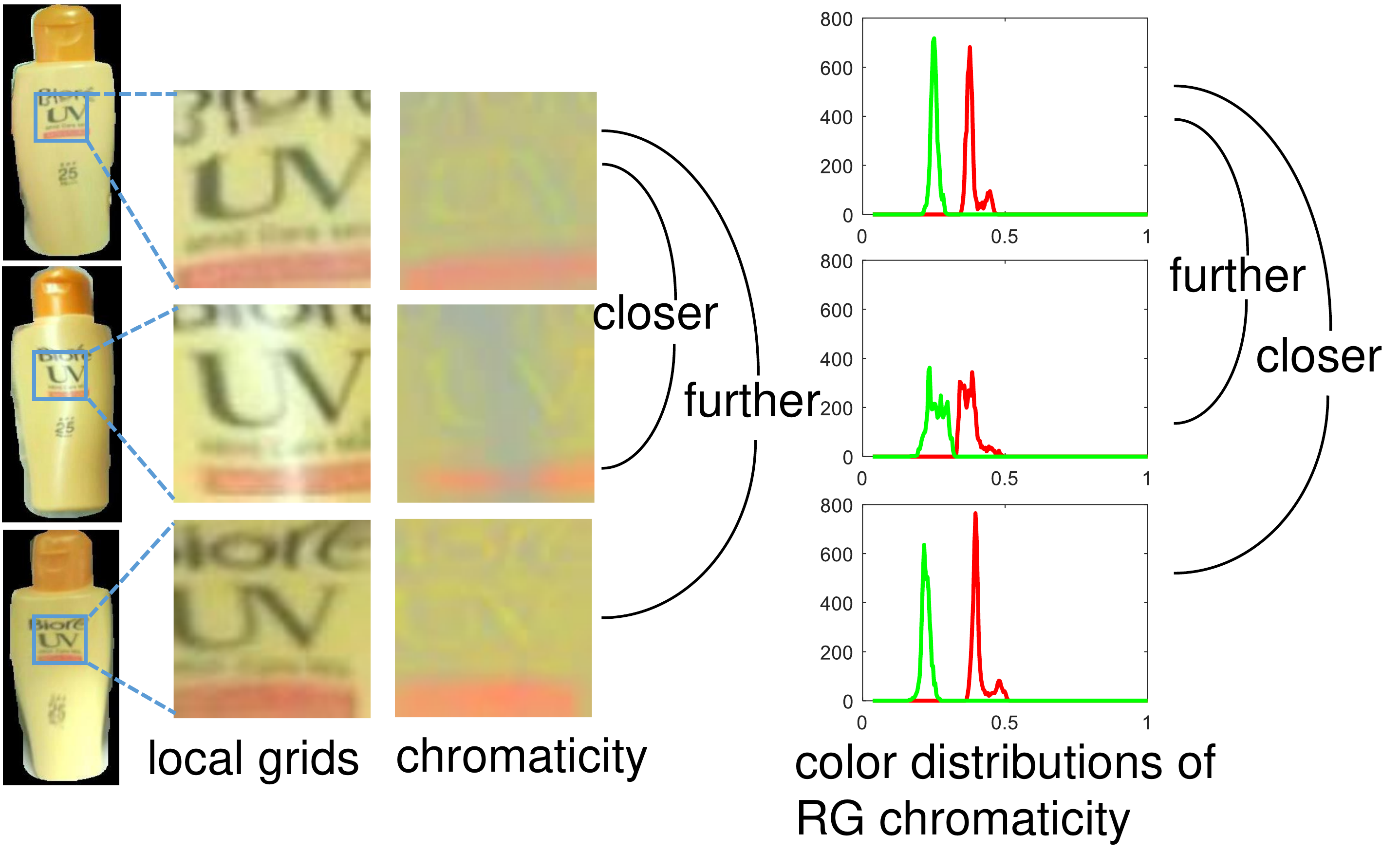}\
    \caption{Distances between color distributions are more sensitive to the presence of highlights than to pixel-to-pixel distance between misaligned images. The grid cells in the top two images are spatially closer to each other, but have greater difference in color distribution due to highlights.}
    \label{fig:colorhist}
\end{figure}

This vector representation of locally re-ordered pixel values is robust to slight misalignment for the following reasons: (1) Since the dimensions of grid cells are much larger than typical misalignment distances, the corresponding grid cells of different images will largely overlap the same object regions; (2) Products tend to have a sparse set of surface colors, and the pixel reordering will help to align these colors between the corresponding grid cells of different images, which is sufficient for measuring color-based consistency; (3) With this representation, the SILR loss is empirically found to be more sensitive to the presence of highlights or albedo distortions than to slight misalignment, as illustrated in Figure~\ref{fig:colorhist} for diffuse chromaticity.

We note that a local color distribution could more directly be modeled by a color histogram. However, color histograms are not differentiable, and this motivated us to develop the pixel reordering representation as a differentiable approximation to color histograms. Local regions that have similar color histograms will have similar pixel reordering representations, and vice versa.

\subsection{Joint finetuning by contrastive loss}\label{allfinetuning}

After training H-Net and S-Net individually, we adopt a novel contrastive loss to finetune the entire network in an end-to-end manner. Our approach is based on the observation that intrinsic image decomposition can be better performed after highlights have been separated from input images. Related observations have been made in other recent works. For example, Ma et al.~\cite{ma2018single} mention that their method cannot handle specularity well, and this limitation will be addressed in future work. Also, Shi et al.~\cite{shi2017learning} discuss that the multiplicative intrinsic image decomposition model, $I_d = A\cdot S$, cannot adequately account for additive highlight components.

\begin{figure}
\centering
\includegraphics[width= \linewidth]{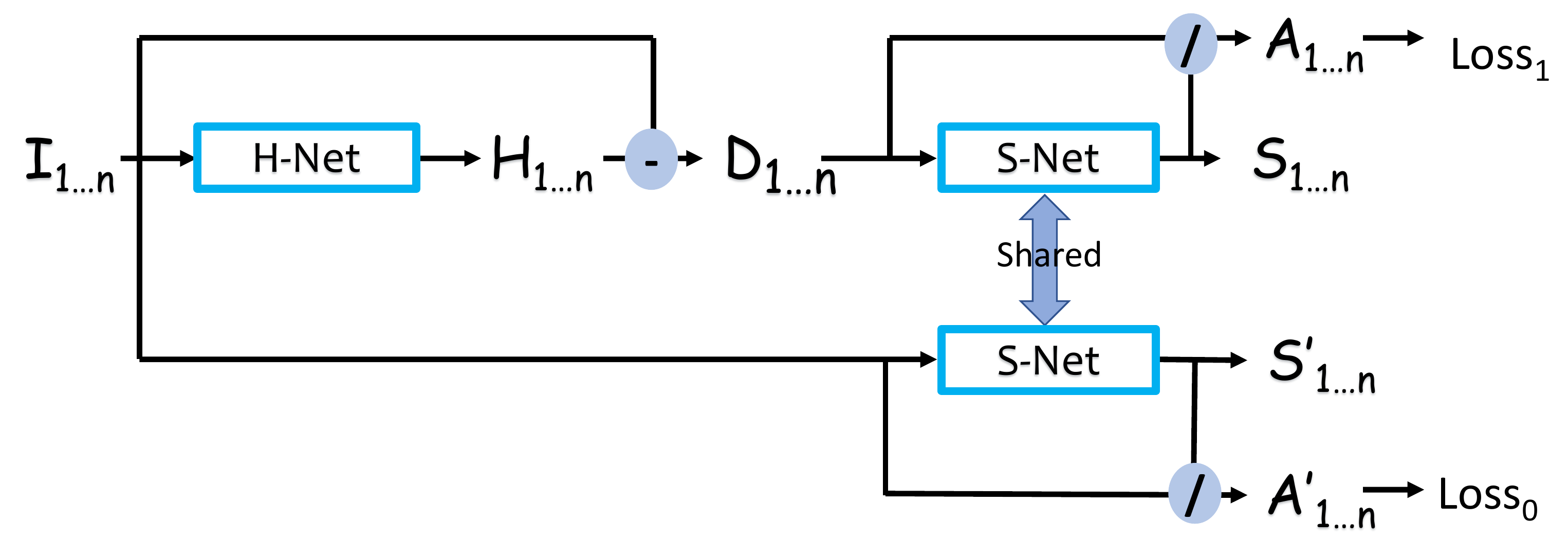}\
   \caption{Network structure for joint finetuning by contrastive loss.
}
\label{fig:contrastive}
\end{figure}

Based on this observation, we define a contrastive loss. As indicated in Figure~\ref{fig:contrastive}, our low-rank loss on the albedo layers of multiple images is $\mathcal{L}_1$ if highlights are removed from the input images following the image formation model $I=A\cdot S+H$. In another branch, we compute the low-rank loss on albedo layers as $\mathcal{L}_0$, where the input images are decomposed by S-Net directly following the image formation model $I=A\cdot S$. The contrastive loss is defined as:
\begin{equation}
\centering
\mathcal{L}_{ct} = \mathcal{L}_1-\mathcal{L}_0.
\label{equation:ctloss}
\end{equation}

Intuitively, the contrastive loss is designed to maximize the distance between $\mathcal{L}_1$ and $\mathcal{L}_0$ (where $\mathcal{L}_{ct}$ is negative), so as to force H-Net to improve its highlight separation and thus decrease $\mathcal{L}_1$ relative to $\mathcal{L}_0$. Both subnets can be finetuned by this loss. In our experiments, we found that using $\mathcal{L}_{ct}$ alone will lead to increases of both $\mathcal{L}_1$ and $\mathcal{L}_0$, as this increases their difference as well. To avoid this degenerate case, we add $\omega \mathcal{L}_1$ as a regularization, such that the joint finetuning loss becomes $\mathcal{L}=\mathcal{L}_{ct}+\omega \mathcal{L}_1$, where $\omega$ is set to 1.0 in our implementation. This ensures that both $\mathcal{L}_1$ and the contrastive loss are minimized together.


After these three training phases, our network shown in Figure~\ref{fig:pipeline} is able to separate the highlight, diffuse, albedo, and shading layers of a test image. Further implementation details are given in the supplement.

%% file: experiment.tex
\section{Experiments}\label{evalutions}

Since previous works generally address highlight separation or intrinsic image estimation but not both, we evaluate our method on these two tasks separately on various datasets. 
Due to limited space, many additional results and analyses, 
including evaluations on the MIT intrinsic image dataset \cite{grosse2009ground} and Intrinsic Images in the Wild (IIW) \cite{bell2014intrinsic}, highlight separation on grayscale images (which cannot be handled by most previous techniques), and the inadequacy of structure-from-motion for aligning our customer photos, are provided in the supplement.


\subsection{Highlight separation}


\subsubsection{Synthetic dataset}\label{comparesynthetic}

In Table~\ref{table:real} (top-left), we compare our method to several leading techniques on highlight separation using synthetic data from the ShapeNet Intrinsic Dataset~\cite{shi2017learning}. From this dataset, we randomly select 500 images covering a wide range of objects and materials to form the test set.  The results are reported in terms of MSE and DSSIM, which measure pixelwise difference and structural dissimilarities, respectively.
\begin{figure}
\centering
\includegraphics[width= \linewidth]{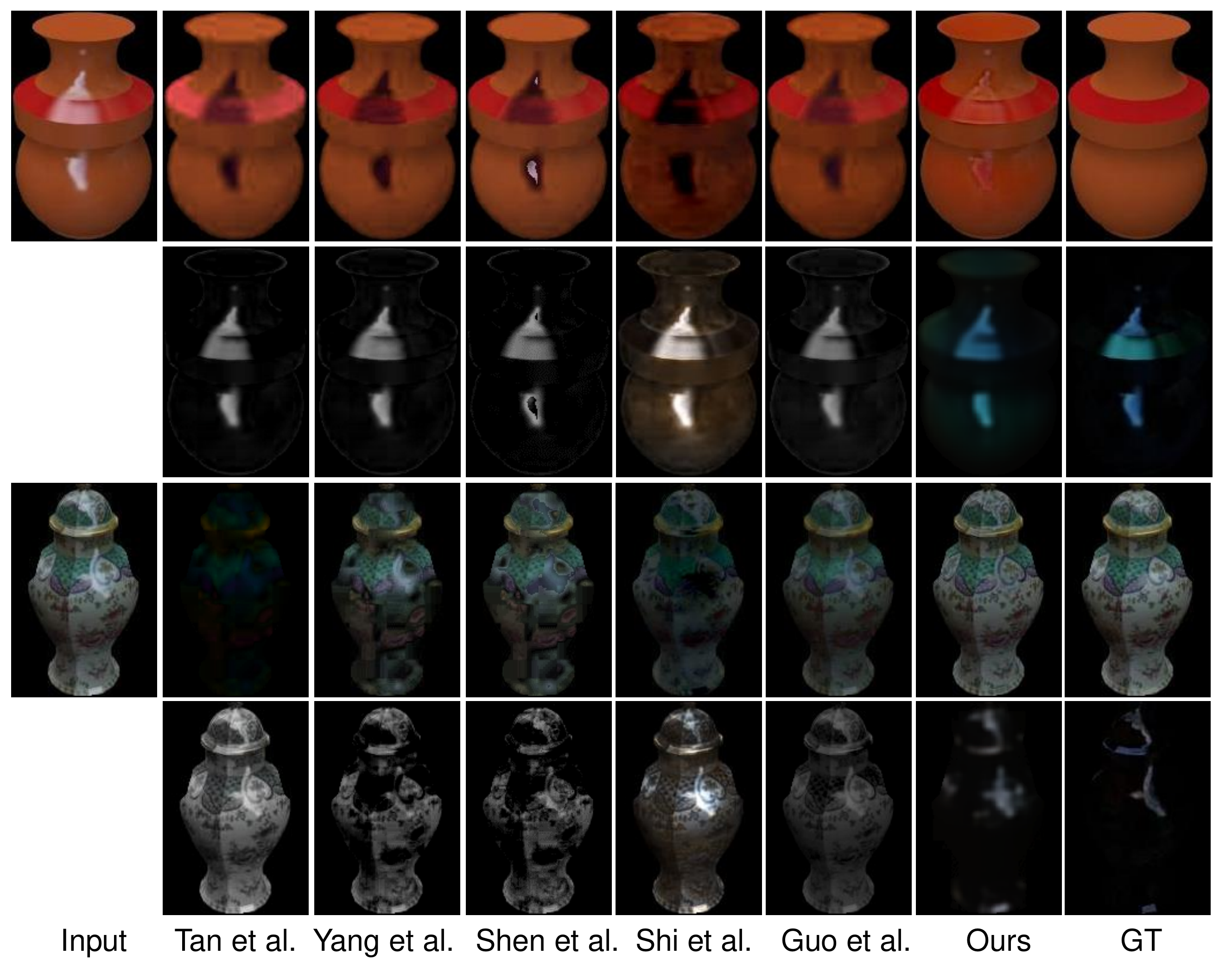}
   \caption{Visual comparisons of highlight separation on the ShapeNet Intrinsic Dataset. For each example, the top row shows the input image and separated diffuse layers, and the bottom row exhibits the separated highlight layers. GT denotes ground truth.}
\label{fig:highlightsynthetic}
\end{figure}

Examples for visual comparison are shown in Figure~\ref{fig:highlightsynthetic}. Earlier methods~\cite{tan2004separating,yang2010real,shen2013real} often assume the illumination to be white and can estimate only a grayscale highlight layer, even when the lighting is not white. Moreover, they cannot deal well with saturated regions, which generally have non-white highlight components that result from subtracting (non-white) diffuse components from saturated image values. A recent method~\cite{guo2018single} handles saturated highlight regions better with a low-rank and sparse decomposition. However, it still cannot recover correct diffuse color at saturated regions where its assumed dichromatic model is violated, leading to artifacts in diffuse layers. The CNN-based method of \cite{shi2017learning} can learn from various training data composed of different surface materials, but it still does not handle saturation well. By comparison, our method succeeds in predicting highlight colors and generates reasonable diffuse layers even for saturated regions. 

\subsubsection{Real dataset}
Since no standard real-image dataset exists for evaluating highlight separation, we captured a dataset consisting of 20 ordinary objects with ground truth obtained by cross polarization in a laboratory environment. Table~\ref{table:real} (top-right) shows the MSE and DSSIM of different methods on this dataset. 
Qualitative comparisons are shown in Figure~\ref{fig:highlightrealimages} and Figure~\ref{fig:realhighlight1} of the supplement.
Our method is found to recover highlight and diffuse layers closest to the ground truth, with highlights of correct color even in saturated regions. While our technique successfully estimates the surface colors in the diffuse layers, the other methods tend to leave black artifacts at saturated regions. Additional qualitative results on real images under natural lighting can be found in the supplement as well. 

\begin{table}[t]
\centering 
\resizebox{.92\columnwidth}{!}{
\begin{tabular}{cc cc ccc} 
\hline 
&&\multicolumn{2}{c}{Synthetic}&&\multicolumn{2}{c}{Real}\\
\cline{3-4}\cline{6-7}
Method&&MSE&DSSIM&&MSE&DSSIM\\
\hline
Tan et al.&&0.0155&0.0616&&0.0173&0.0368\\
Yang et al.&&0.0053&0.0336&&0.0043&0.0162\\
Shen et al.&&0.0059&0.0338&&0.0047&0.0163\\
Shi et al.&&0.0063&0.0526&&0.0063&0.0237\\
Guo et al.&&0.0028*&0.0208*&&0.0045&\textcolor{blue}{0.0145}\\
Ours&&\textcolor{blue}{0.0016}&\textcolor{red}{0.0159}&&\textcolor{red}{0.0036}&\textcolor{red}{0.0139}\\
\hline
No Finetuning&&\textcolor{red}{0.0015}&0.0176&&0.0045&0.0188\\
Pixel-to-pixel&&0.0020&\textcolor{blue}{0.0166}&&\textcolor{blue}{0.0041}&0.0149\\
\hline
\end{tabular}
}
\caption{Highlight separation on the synthetic ShapeNet Intrinsic Dataset and on a real-image dataset. Errors are for diffuse layers. {\bf Top:} Comparison to state of the art. Lowest errors shown in red, and second lowest in blue. Guo\cite{guo2018single} is tested on only 50 of the 500 synthetic data in total, with the results marked by *, since we needed the authors to process our images. {\bf Bottom:} Ablations.\label{table:real}}
\end{table}

\subsubsection{Ablations}

We conducted an ablation study to examine the main novel elements of our system, with the results shown in Table~\ref{table:real} (bottom). When the unsupervised finetuning is removed from the system, the difference in performance becomes more significant on real images than on synthetic images, since the finetuning provides training in the domain of real images. On real images without finetuning, the performance is at a level similar to the previous state of the art, while our full system yields an approximate 20-25\% improvement over this.

To examine the importance of our color distribution loss in dealing with misalignment, we compare to the results of our network when using a pixel-to-pixel low-rank loss instead. Some moderate quantitative gain is observed, about 4-20\% for synthetic images and 7-10\% for real images. We point readers to the qualitative comparisons shown in Figure~\ref{fig:misalignrobust} of the supplement, where the diffuse layers computed without the color distribution loss contain severe artifacts around highlight regions. Later, it will be shown that the color distribution loss has greater quantitative impact on intrinsic image decomposition.

When the contrastive loss is removed from the system, the solution often degenerates to a diffuse layer of all zeros, as this allows H-Net to reach a minimum most quickly. Similar to a generative adversarial network (GAN), the contrastive loss creates a competition between losses that can steer the learning toward better minima and/or away from degenerate cases. By including the contrastive loss, the learning rate of S-Net becomes twice that of H-Net, causing the training to focus more on S-Net and thus avoiding degenerate solutions.


\begin{table}[t]
\centering 
\resizebox{.93\columnwidth}{!}{
\begin{tabular}{cccccc} 
\hline 
&&MSE(A)&DSSIM(A)&MSE(S)&DSSIM(S)\\
\hline
SIRFS&&0.0081&0.0636&0.0066&0.0785\\
DI&&0.0086&0.0590&0.0047&0.0765\\
Shi et al.&&0.0068&0.0565&\textcolor{red}{0.0023}&\textcolor{blue}{0.0691}\\
Li et al.&&\textcolor{blue}{0.0066}&0.0541&0.0063&0.0812\\
Ours&&\textcolor{red}{0.0054}&\textcolor{red}{0.0436}&\textcolor{blue}{0.0045}&\textcolor{red}{0.0686}\\
\hline
No Finetuning&&0.0108&0.0664&0.0096&0.0810\\ 
Pixel-to-pixel&&0.0067&\textcolor{blue}{0.0460}&0.0087&0.0774\\ 
\hline
\end{tabular}
}
\caption{Intrinsic image decomposition on synthetic data from the ShapeNet Intrinsic Dataset. The lowest errors are highlighted in red and the second lowest are in blue. \label{table:intrinsicsynthetic}} 
\end{table}

\subsection{Intrinsic image decomposition}

\subsubsection{ShapeNet Intrinsic Dataset}

For intrinsic image decomposition, we compare our network to SIRFS~\cite{barron2015shape}, DI~\cite{narihira2015direct}, Shi et al.~\cite{shi2017learning}, and Li et al.~\cite{li2018learning} on the ShapeNet Intrinsic Dataset.
Similar to the evaluation of highlight separation, we use MSE and DSSIM to measure results. 
These results are summarized in Table~\ref{table:intrinsicsynthetic} (top) and show the relatively strong performance of our method. Qualitative comparisons are shown in Figure~\ref{fig:shapenetintrinsic} and Figure~\ref{fig:syntheticintrinsicsupple} of the supplement.

SIRFS~\cite{barron2015shape}, which is based on scene priors, fails on non-Lambertian objects. The learning-based method DI~\cite{narihira2015direct} trained on synthetic diffuse scenes exhibits similar problems. 
The method by Shi et al.~\cite{shi2017learning} performs better than previous methods on non-Lambertian objects. One reason is that, like our method, it explicitly models highlights, in contrast to other methods~\cite{narihira2015direct,barron2015shape,li2018learning} which consequently have artifacts in the albedo layer on highlight regions. Another reason is because it is trained on the ShapeNet Intrinsic training split with 80\% of the whole dataset.
In comparison, our method is pretrained on a very small amount (1.1\%) of the ShapeNet dataset to obtain a good network initialization, and is finetuned on a large amount of real data. Despite this, it still performs well on synthetic ShapeNet images. 
Since our S-Net solves for shading and then computes albedo using the image formation model $I_d=A\cdot S$, it generates high resolution albedo maps with texture details, whereas many networks that directly solve for albedo will obtain blurred results due to feature map downsampling in the network.  

\subsubsection{DiLiGenT dataset}

We also conduct experiments on real images. Since there do not exist intrinsic image datasets with ground truth for general real objects\footnote{The IIW dataset~\cite{bell2014intrinsic} and SAW dataset~\cite{kovacs2017shading} are of real {\em scenes}, while the objects in the MIT dataset~\cite{grosse2009ground} are restricted to highly Lambertian reflectance.}, we evaluate on ground-truth shading layers generated from the DiLiGenT photometric stereo dataset~\cite{shi2019diligent}. As DiLiGenT provides ground-truth surface normals and lighting, but no reflectance information, only the shading layers can be reconstructed. The dataset contains images of 10 non-Lambertian objects under 96 different lighting conditions.

\begin{figure}
\centering
\includegraphics[width= \linewidth]{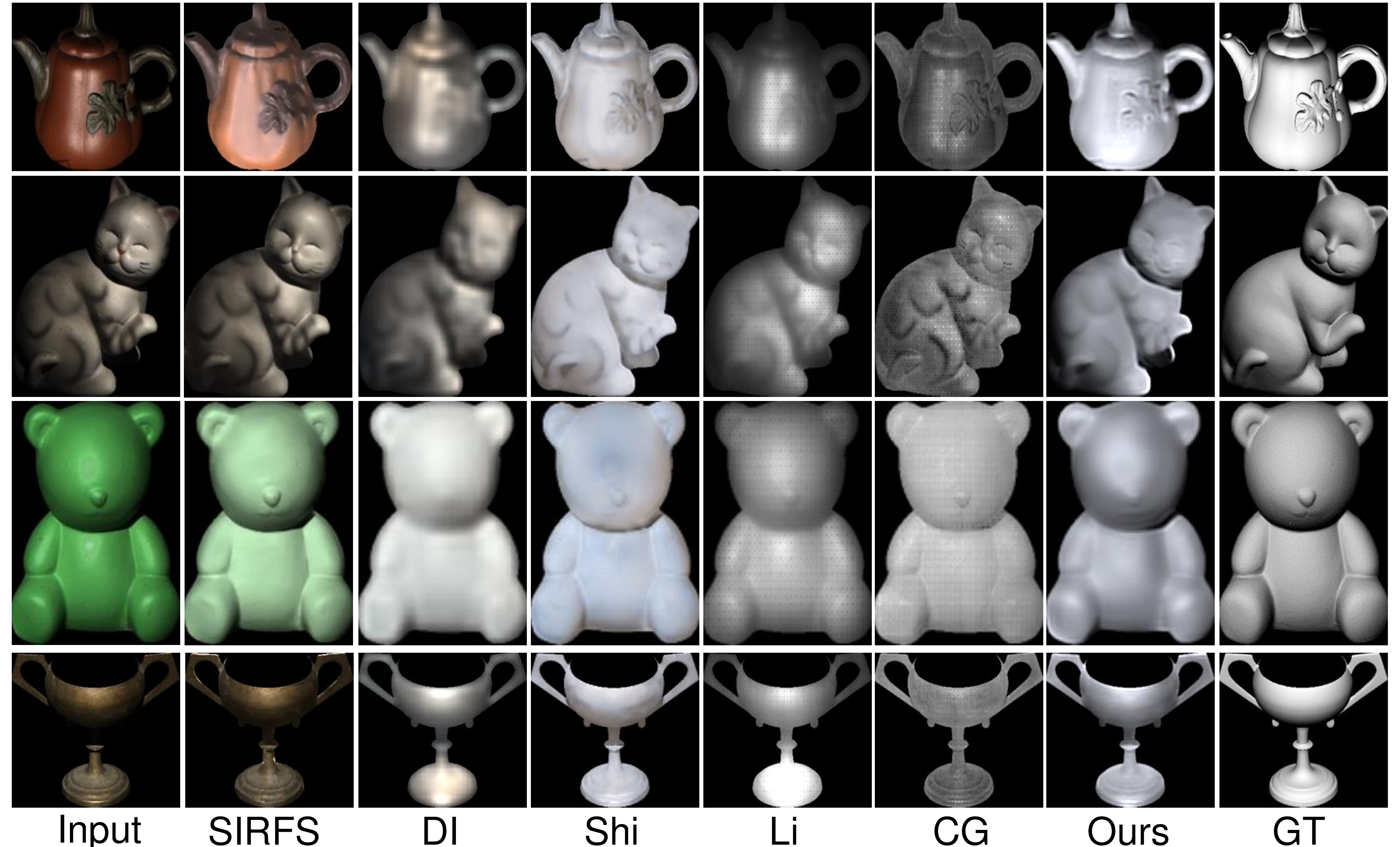}\
   \caption{Shading layer comparisons on DiLiGenT dataset.
   Please see Table~\ref{table:diligent} for the notations of previous methods.
   }
\label{fig:diligent}
\end{figure}

Comparisons of our network are made to several leading techniques.
Qualitative and quantitative results are shown in Figure~\ref{fig:diligent} and Table~\ref{table:diligent}. It is found that our network yields the highest accuracy in this challenging case of real non-Lambertian objects.


\begin{table}[t]
\centering 
\scalebox{0.8}{
\begin{tabular}{cccc} 
\hline 
&&MSE(S)&DSSIM(S)\\
\hline
SIRFS\cite{barron2015shape}&&0.0097&0.0457\\
DI\cite{narihira2015direct}&&0.0061&0.0385\\
Shi\cite{shi2017learning}&&0.0043&0.0331\\
Li\cite{li2018learning}&&0.0073&0.0401\\
CG\cite{li2018cgintrinsics}&&0.0061&0.0413\\
Ours&&\textcolor{red}{0.0041}&\textcolor{red}{0.0316}\\
\hline
\end{tabular}
}
\caption{Evaluation of shading accuracy on the DiLiGenT dataset. The lowest errors are highlighted in red. } \label{table:diligent}
\end{table}

\subsubsection{Other datasets}
\begin{figure}
\centering
\includegraphics[width= \linewidth]{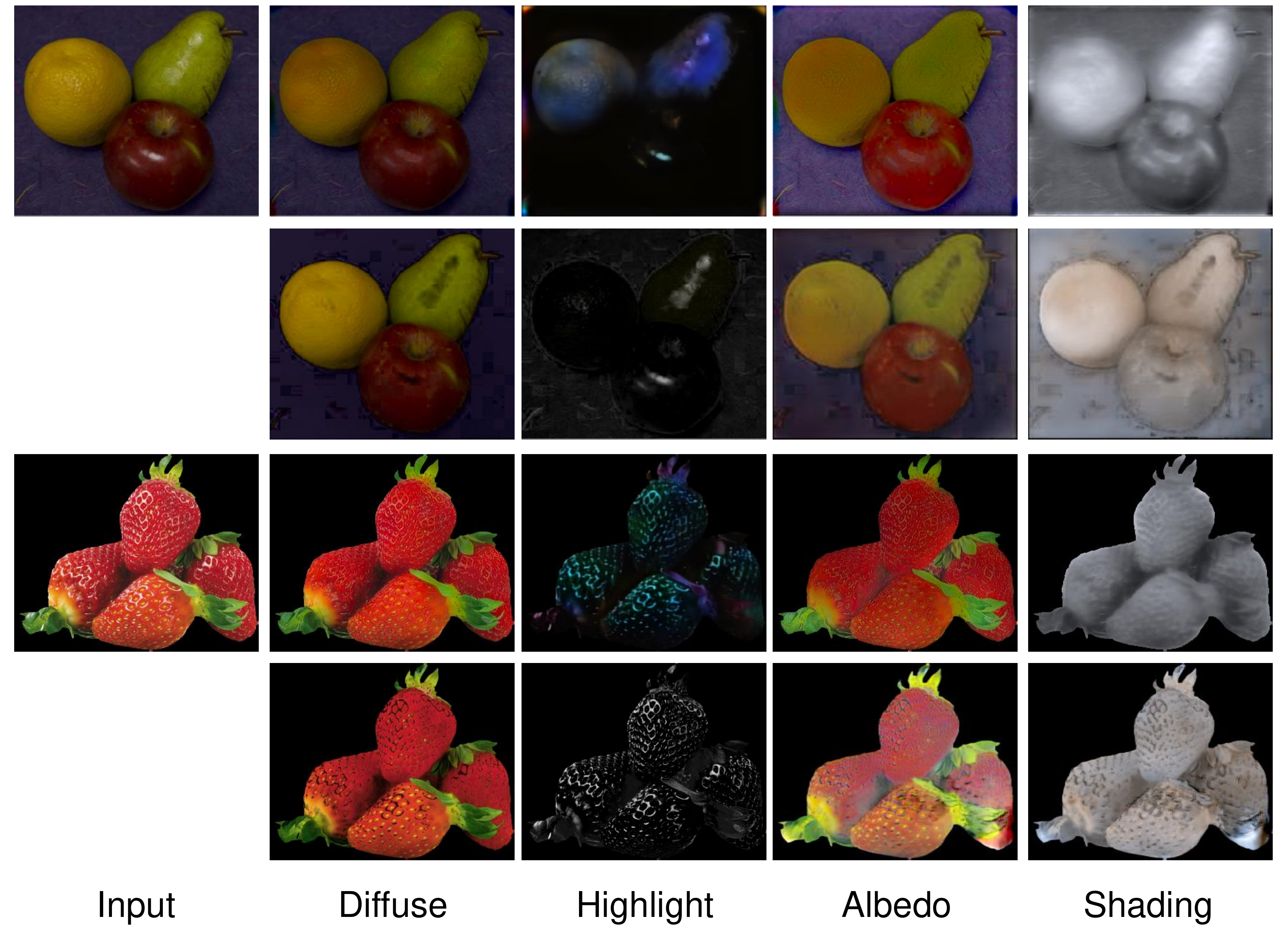}\
   \caption{Qualitative comparisons on real images. We compare our end-to-end separation (odd rows) to the combination of Yang~\cite{yang2010real} for highlight separation and Shi~\cite{shi2017learning} for intrinsic image decomposition (even rows).
}
\label{fig:qualitative}
\end{figure}

There exist other datasets that can be used for intrinsic image evaluation, including the MIT intrinsic image dataset~\cite{grosse2009ground} and Intrinsic Images in the Wild (IIW)~\cite{bell2014intrinsic}. Due to limited space, comparisons on these datasets, as well as qualitative comparisons on more natural images collected from the Internet, are presented in the supplement. In addition, some qualitative results of full end-to-end separations on real images are shown in Figure~\ref{fig:qualitative}, with comparisons to a combination of two previous methods that exhibit state-of-the-art performance in quantitative evaluations.



\subsubsection{Ablations}

Ablation experiments were also conducted for intrinsic image decomposition on ShapeNet, with the results given in Table~\ref{table:intrinsicsynthetic} (bottom). Even though ShapeNet consists of synthetic images, significant gains were obtained by including the unsupervised finetuning (15-50\%) and by using the color distribution loss instead of a pixel-to-pixel low rank loss (5-48\%). The difference is particularly large for shading, as also evidenced in the qualitative comparisons shown in Figure~\ref{fig:misalignrobust} of the supplement where the shading layers are more indicative of surface shape. As with highlight separation, removal of the contrastive loss leads to degenerate solutions where the diffuse layer is all zero.

